\documentclass{article}

\usepackage{microtype}
\usepackage{graphicx}
\usepackage{subcaption}
\usepackage{booktabs} % for professional tables
\usepackage{multirow}

\usepackage{hyperref}

\usepackage[preprint]{icml2026}

\usepackage{amsmath}
\usepackage{amssymb}
\usepackage{mathtools}
\usepackage{amsthm}
\usepackage{physics}

\usepackage[capitalize,noabbrev]{cleveref}

\theoremstyle{plain}

\theoremstyle{definition}

\theoremstyle{remark}

\usepackage[textsize=tiny]{todonotes}

\icmltitlerunning{}

\begin{document}

\twocolumn[
  \icmltitle{AbLWR: A Context-Aware Listwise Ranking Framework for Antibody-Antigen Binding Affinity Prediction via Positive-Unlabeled Learning}

  % It is OKAY to include author information, even for blind submissions: the
  % style file will automatically remove it for you unless you've provided
  % the [accepted] option to the icml2026 package.

  % List of affiliations: The first argument should be a (short) identifier you
  % will use later to specify author affiliations Academic affiliations
  % should list Department, University, City, Region, Country Industry
  % affiliations should list Company, City, Region, Country

  % You can specify symbols, otherwise they are numbered in order. Ideally, you
  % should not use this facility. Affiliations will be numbered in order of
  % appearance and this is the preferred way.
  \icmlsetsymbol{equal}{*}

  \begin{icmlauthorlist}
    \icmlauthor{Fan Xu}{ad1}
    \icmlauthor{Zhi-an Huang}{ad2}
    \icmlauthor{Haohuai He}{ad1}
    \icmlauthor{Yidong Song}{ad3}
    \icmlauthor{Wei Liu}{ad4}
    \icmlauthor{Dongxu Zhang}{ad4}
    \icmlauthor{Yao Hu}{ad1}
    \icmlauthor{Kay Chen Tan}{ad1}
  \end{icmlauthorlist}

  \icmlaffiliation{ad1}{Department of Data Science and Artificial Intelligence, The Hong Kong Polytechnic University, Hong Kong SAR, China}
  \icmlaffiliation{ad2}{Department of Computer Science, City University of Hong Kong (Dongguan), Dongguan, China}
  \icmlaffiliation{ad3}{School of Computer Science and Engineering, Sun Yat-sen University, Guangzhou, Guangdong, China}
  \icmlaffiliation{ad4}{School of Public Health, Xiamen University, Xiamen, China}

  \icmlcorrespondingauthor{Zhi-an Huang}{huang.za@cityu-dg.edu.cn}
  \icmlcorrespondingauthor{Kay Chen Tan}{kctan@polyu.edu.hk}

  % You may provide any keywords that you find helpful for describing your
  % paper; these are used to populate the "keywords" metadata in the PDF but
  % will not be shown in the document
  \icmlkeywords{Machine Learning, ICML}

  \vskip 0.3in
]

% this must go after the closing bracket ] following \twocolumn[ ...

% This command actually creates the footnote in the first column listing the
% affiliations and the copyright notice. The command takes one argument, which
% is text to display at the start of the footnote. The \icmlEqualContribution
% command is standard text for equal contribution. Remove it (just {}) if you
% do not need this facility.

% Use ONE of the following lines. DO NOT remove the command.
% If you have no special notice, KEEP empty braces:
\printAffiliationsAndNotice{}  % no special notice (required even if empty)
% Or, if applicable, use the standard equal contribution text:
% \printAffiliationsAndNotice{\icmlEqualContribution}

\begin{abstract}
Accurate prediction of antibody-antigen binding affinity is fundamental to therapeutic design, yet remains constrained by severe label sparsity and the complexity of antigenic variations. In this paper, we propose AbLWR (Antibody-antigen binding affinity List-Wise Ranking), a novel framework that reformulates the conventional affinity regression task as a listwise ranking problem. To mitigate label sparsity, AbLWR incorporates a PU (Positive-Unlabeled) learning mechanism leveraging a dual-level contrastive objective and meta-optimized label refinement to learn robust representations. Furthermore, we address antigenic variation by employing a homologous antigen sampling strategy where Multi-Head Self-Attention (MHSA) explicitly models inter-sample relationships within training lists to capture subtle affinity nuances. Extensive experiments demonstrate that AbLWR significantly outperforms state-of-the-art baselines, improving the Precision@1 (P@1) by over 10$\%$ in randomized cross-validation experiments. Notably, case studies on Influenza and IL-33 validate its practical utility, demonstrating robust ranking consistency in distinguishing subtle viral mutations and efficiently prioritizing top-tier candidates for wet-lab screening.

\end{abstract}

\section{Introduction}
% The introduction of antibody and the binding affinity
Antibodies are central to the adaptive immune response, neutralizing pathogens via specific antigen binding mediated by complementarity-determining regions (CDRs) \cite{jin2022emerging, kim2023computational}. The potency and efficacy of these interactions are typically evaluated using quantitative metrics, such as the dissociation constant ($K_{\text{d}}$), half-maximal inhibitory or effective concentrations ($\text{IC}{50}$/$\text{EC}{50}$), and escape fractions. Currently, the rapid evolution of antigens poses a significant challenge to antibody effectiveness \cite{hie2021learning, yang2023antigen, sun2025novel}, creating a continually shifting antigenic landscape where emerging viral strains can evade the neutralizing effects of therapeutic antibodies \cite{zhang2021membrane, garcia2021multiple}. Thus, developing methods to accurately identify and prioritize the most potent antibodies for specific, evolving antigens is of paramount importance.

% However, the binding affinity is difficult to be measured. A lot of work has been done for this task.
Conventionally, binding affinity metrics are quantified through wet-lab experiments, such as Surface Plasmon Resonance (SPR), Enzyme-Linked Immunosorbent Assays (ELISA), and pseudovirus neutralization assays \cite{wang2022development}. However, these procedures are resource-intensive and time-consuming, particularly when antigenic drift necessitates multiple repeated experiments. Consequently, significant computational efforts have been directed toward binding affinity prediction. Existing methodologies can be broadly classified into four paradigms, spanning from traditional physics-based approaches to modern data-driven deep learning pipelines (sequence-, structure-, and complex-based) \cite{frey2025lab}. Sequence-based models leverage protein language models (PLMs) to capture antibody-specific features directly from primary sequences \cite{ruffolo2021deciphering, ullanat2026learning}. To exploit geometric priors, structure-based methods \cite{nguyen2021graphdta, liu2025abrank} integrate structural information, often utilizing predicted coordinates from the advanced structure prediction tools \cite{abramson2024accurate, ruffolo2023fast, wohlwend2025boltz}. Targeting the bound state, complex-based approaches explicitly model the 3D conformation of the bound state \cite{michalewicz2024antipasti}. Distinct from these learning-based frameworks, traditional methods rely on empirical force fields and physics-based energy functions for stability estimation \cite{schymkowitz2005foldx}.

\begin{figure}[htb]
  \vskip 0.2in
  \begin{center}
    \centerline{\includegraphics[width=\columnwidth, trim=0 0 0 0, clip]{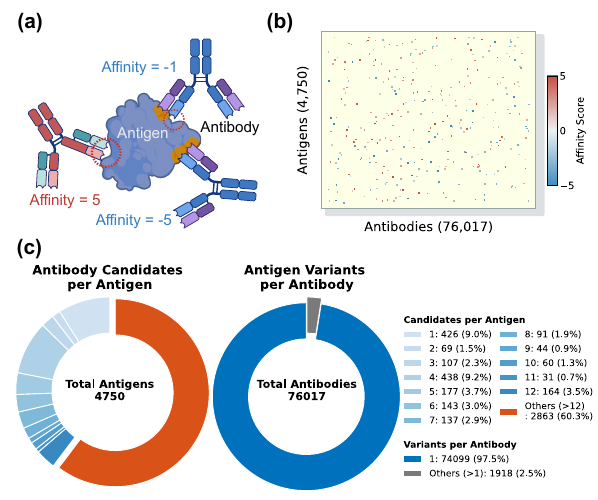}}
    \caption{\textbf{Key challenges in Ab-Ag binding affinity prediction.} (a) Distinct affinity profiles of multiple antibodies binding to the same antigen. (b) The large reservoir of unlabeled data remains largely unexploited by existing supervised paradigms, compared with the immense volume of antibody and antigen sequences. (c) Analysis of existing binding affinity data from public datasets (refer to Appendix \cref{Data Collection} for data collection details) reveals that most antigens are associated with different antibody candidates, whereas individual antibodies typically bind to a single specific antigen.}
    \label{fig1}
  \end{center}
\end{figure}

% However, the results of prediction are limitted by two problems: 1) similar ag-ab pairs, 2) label sparsity 
Despite these advancements, the predictive performance remains limited by two critical bottlenecks. First, existing models, which predominantly rely on point-wise regression, struggle to discriminate affinity levels among highly homologous Antibody-Antigen (Ab-Ag) pairs \cite{mason2021optimization}. As illustrated in \cref{fig1} (a) and (c), multiple antibodies can bind to the identical antigen yet exhibit distinct affinity profiles. Treating these samples in isolation often fails to capture such subtle discrepancies in their relative binding strength. While pairwise ranking approaches \cite{liu2025abrank} attempt to mitigate this, they are restricted to local comparisons, lacking the global context required to accurately prioritize top candidates. Second, the field suffers from severe label sparsity. While sequence data is immense, experimentally verified affinity labels are scarce (\cref{fig1} (b)) \cite{dunbar2014sabdab, jankauskaite2019skempi}. Crucially, current fully supervised approaches largely overlook the vast potential of unlabeled Ab-Ag pairs, leaving a rich source of structural information unexploited.

% In order to solve this problem, our method...
Thus, we propose AbLWR, a novel two-stage framework that reformulates binding affinity prediction as a listwise ranking task. Rather than regressing absolute values, AbLWR aims to learn the relative order of Ab-Ag pairs based on binding strength. In the initial pre-training stage, we leverage Positive-Unlabeled (PU) learning to harness the abundance of unlabeled data for learning intrinsic structural representations. By treating experimentally verified pairs as positives and unverified pairs as unlabeled data, this module trains a classifier to stratify binding affinities into coarse-grained levels. In the subsequent ranking stage, we construct training lists comprising Ab-Ag pairs associated with identical or homologous antigens. To capture subtle affinity nuances, we develop a context-aware ranking module incorporating Multi-Head Self-Attention (MHSA) mechanism, which explicitly models the inter-dependencies among candidate pairs.

Our contributions are summarized as follows: 
\begin{itemize}
    \item We propose AbLWR, the first attempt to reformulate affinity prediction from point-wise regression to listwise ranking. This shift aligns the optimization objective with candidate prioritization, capturing the relative affinity manifold more effectively than regression baselines.
    \item We introduce a semi-supervised PU learning strategy to tackle label sparsity. By distilling interaction priors from unlabeled Ab-Ag pairs, this approach yields robust representations that generalize well beyond limited experimental data.
    \item We devise a context-aware architecture incorporating MHSA module to model the global dependencies among candidate pairs. This mechanism enables the model to perform contrastive reasoning, directly distinguishing subtle residue variations across homologous sequences.
    \item Extensive evaluations on independent testing sets and task-specific splits demonstrate that AbLWR achieves state-of-the-art (SOTA) performance. Our results highlight the framework's superior ability to identify high-affinity variants in lead optimization scenarios.
\end{itemize}

\section{Related Work}
\subsection{Binding Affinity Prediction} \label{RelatedWork - Binding}
Binding affinity is a critical determinant of therapeutic efficacy, governed by intricate physicochemical factors \cite{akbar2022progress}. Historically, prediction relied on physics-based simulations. Tools like Rosetta \cite{chaudhury2010pyrosetta} and FoldX \cite{schymkowitz2005foldx} employ empirical force fields to estimate Gibbs free energy changes ($\Delta\Delta G$) from the bound complex. While interpretable, these methods are computationally intensive and sensitive to structural noise. The expansion of datasets like SAbDab \cite{dunbar2014sabdab} and SKEMPI \cite{jankauskaite2019skempi} has accelerated a shift toward data-driven approaches. In this review, we categorize these deep learning methodologies based on their input modalities into sequence-, structure-, and complex-based methods.

\textbf{Sequence-based methods.} These approaches model affinity directly from amino acid sequences. Early efforts utilized 1D convolutional neural networks (CNNs) or recurrent neural networks (RNNs) to extract local motifs \cite{liberis2018parapred, mason2021optimization, huang2022abagintpre}. Recently, PLMs have become dominant: general-purpose models like ESM-1b \cite{rives2021biological} and antibody-specific ones like AntiBERTy \cite{ruffolo2021deciphering} and MINT \cite{ullanat2026learning} leverage Masked Language Modeling to generate embeddings encapsulating evolutionary semantics. These representations are widely used for regression tasks \cite{gao2023pre}. However, these methods inherently lack explicit spatial context, limiting their efficacy in capturing long-range interactions and discontinuous epitopes. 

\textbf{Structure-based methods.} To address the limitations of sequence-only models, structure-based methods incorporate geometric data to enhance accuracy, typically deriving information from separate antibody or antigen structures. These approaches often employ geometric deep learning architectures to capture local spatial features, such as 3D-CNNs \cite{wang2020topology} and graph neural networks (GNNs) \cite{bandara2025deep}. Furthermore, the integration of advanced structure predictors like AlphaFold-3 \cite{abramson2024accurate}, IgFold \cite{ruffolo2023fast}, and Boltz-1 \cite{wohlwend2025boltz} enables these models to utilize high-quality predicted structures. 

\textbf{Complex-based methods.} Deep learning complex-based methods operate directly on the 3D conformation of the bound state to capture interaction details. Recently, emerging methods like ANTIPASTI \cite{michalewicz2024antipasti} have been proposed to interpret complex stability. Despite their potential precision, these data-driven methods depend heavily on high-quality co-crystal structures or docked poses, which are often scarce and difficult to obtain for novel variants.

In spite of these advances, current data-driven approaches are mainly constrained by the formulation of affinity prediction as a direct regression problem. This often fails to discriminate between homologous variants with subtle mutations. Our work addresses this by shifting the paradigm to a listwise ranking task, explicitly modeling the relative order of candidates.

\subsection{PU Learning}  \label{RelatedWork - PUL}
Standard supervised learning relies on fully labeled datasets containing both positive and negative instances. However, in many biochemical scenarios, obtaining confirmed negative samples is experimentally expensive or infeasible, resulting in datasets composed solely of verified positives and a large pool of unlabeled data. PU learning addresses this challenge by training a classifier using only the positive and unlabeled sets \cite{bekker2020learning}.

Existing PU learning methods generally fall into two categories: sample selection and cost-sensitive learning. Sample selection approaches first identify reliable negative samples from the unlabeled set. Early methods relied on heuristics \cite{chaudhari2012learning} or auxiliary models \cite{luo2021pulns, hu2021predictive, li2022your}, while recent advances have incorporated recursive sampling strategies \cite{xinrui2023beyond} and consistency regularization \cite{wang2024contrastive} to improve the robustness of negative identification. Although intuitive, these methods heavily depend on the quality of the initial selection and have the risk of discarding informative hard-negative samples. In contrast, cost-sensitive learning methods treat unlabeled data as a mixture of positives and negatives, formulating an unbiased risk estimator to train the model directly. Following the foundational convex formulations \cite{du2015convex} and non-negative PU losses \cite{kiryo2017positive}, recent research has focused on relaxing the strict assumptions of known class priors and random data selection \cite{long2024positive, elkan2008learning}. Notably, novel frameworks have been proposed to handle arbitrary positive shifts \cite{wang2023pue}, extreme class imbalance \cite{fu2024integrated}, and prior-free estimation via graph cuts \cite{zhao2023class}, extending PU learning to more complex and realistic scenarios. Motivated by these advances, we leverage the capability of PU learning to mine latent interaction patterns from massive unlabeled data, thereby addressing the label sparsity in affinity prediction.

\begin{figure*}[htb]
  \vskip 0.2in
  \begin{center}
    \centerline{\includegraphics[width=\linewidth, trim=0 380 0 0, clip]{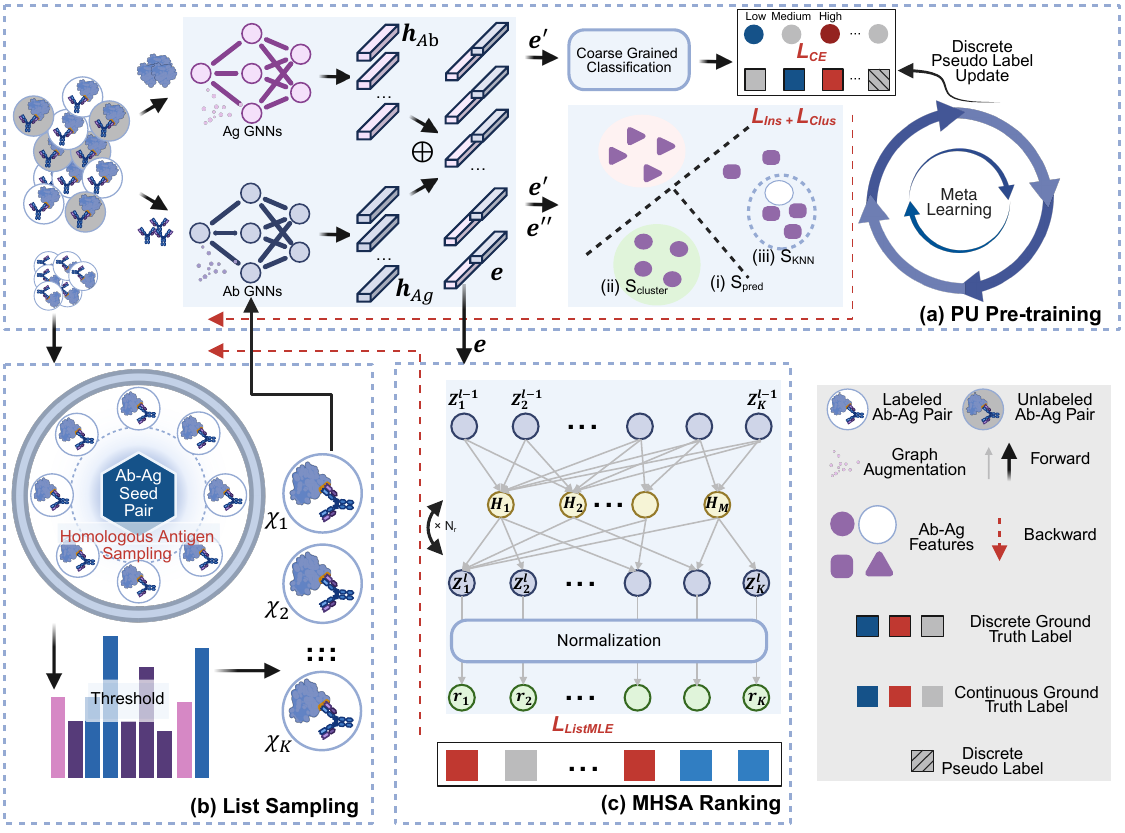}}
    \caption{\textbf{The overall framework of AbLWR.} The architecture comprises three key components: (a) PU Pre-training: Dual GNN encoders learn robust representations from the combined dataset ($\mathcal{D}$) by jointly optimizing a composite contrastive objective ($\mathcal{L}_{\text{Ins}} + \mathcal{L}_{\text{Clus}}$) and a cross entropy loss ($\mathcal{L}_{\text{CE}}$) with meta-reweighted pseudo-labels. (b) List Sampling: A homologous sampling strategy constructs informative training lists $\mathcal{T}$ based on antigen sequence similarity ($\delta_{\text{seq}}$) and affinity margins ($y_{\text{cutoff}}$). (c) MHSA Ranking: The sampled lists are processed by MHSA mechanism to capture intra-list interactions via self-attention, yielding the final affinity ranking scores $\mathbf{r}$.}
    \label{fig2}
  \end{center}
\end{figure*}

\section{Methods}
This section outlines the methodology of AbLWR (see in \cref{fig2}). We begin by establishing the mathematical notations and problem definition in \cref{Problem Definition}. Subsequently, \cref{Model Architecture} details the model architecture. Finally, training details including the learning objectives and the datasets are formulated in \cref{Training}.

\subsection{Problem Definition} \label{Problem Definition}
For a given antigen, we represent its primary sequence as $\mathbf{S}_{\text{Ag}} = (b_1, \dots, b_{L_{\text{Ag}}})$, where $L_{\text{Ag}}$ denotes the sequence length. For the antibody, we focus exclusively on the binding interface, representing it by its six CDRs with length $L_{\text{Ab}}$: $\mathbf{S}_{\text{Ab}} = (a_1, \dots, a_{L_{\text{Ab}}})$. To incorporate geometric constraints, we leverage pre-trained structure prediction models: IgFold \cite{ruffolo2023fast} for antibodies and ESMFold \cite{lin2023evolutionary} for antigens. The resulting 3D structures are represented as point clouds $\mathbf{X}_{\text{Ab}} = \{\mathbf{x}_1, \dots, \mathbf{x}_{L_{\text{Ab}}}\}$ and $\mathbf{X}_{\text{Ag}} = \{\mathbf{x}_1, \dots, \mathbf{x}_{L_{\text{Ag}}}\}$, where $\mathbf{x}_j \in \mathbb{R}^{C \times 3}$ denotes the coordinates of $C$ atoms in the $j$-th residue. Formally, we define a single Ab-Ag pair instance as the tuple: $\mathcal{X} \triangleq (\mathbf{S}_{\text{Ab}}, \mathbf{S}_{\text{Ag}}, \mathbf{X}_{\text{Ab}}, \mathbf{X}_{\text{Ag}})$.

We frame the binding affinity prediction problem as a listwise ranking task. Let $\mathcal{T} = \{ \mathcal{X}_1, \dots, \mathcal{X}_K \}$ denote a query list comprising $K$ Ab-Ag pairs (where $K=5$), and $\mathbf{y} = [y_1, \dots, y_K]^\top$ be the corresponding ground-truth binding affinities. Our goal is to learn a scoring function $f: \mathcal{T} \to \mathbb{R}^K$ that predicts ranking scores $\mathbf{r} = [r_1, \dots, r_K]^\top$ such that the ranking order of $\mathbf{r}$ is consistent with $\mathbf{y}$.

\subsection{Model Architecture} \label{Model Architecture}
\paragraph{Ab-Ag GNN Encoders}
We further model the Ab-Ag pair $\mathcal{X}$ as residue-level molecular graphs, $\mathcal{G}_{\text{Ab}} = (\mathcal{N}_{\text{Ab}}, \mathcal{E}_{\text{Ab}})$ and $\mathcal{G}_{\text{Ag}} = (\mathcal{N}_{\text{Ag}}, \mathcal{E}_{\text{Ag}})$. Edges $(u, v) \in \mathcal{E}$ are established based on the coordinates $\mathbf{X}$ if the Euclidean distance between any non-hydrogen atoms in residues $u$ and $v$ is less than $4.5\,\text{\AA}$ \cite{pittala2020learning} . Initial node features $\mathbf{H}^{(0)}$ are derived by encoding the sequences $\mathbf{S}$ with PLMs: IgFold for antibodies and ESM-2 for antigens.

To independently model the distinct structural topologies of the antibody and antigen, we employ dual Graph Convolutional Network (GCN) encoders, $\phi_{\text{Ab}}$ and $\phi_{\text{Ag}}$ \citep{kipf2016semi}. Both encoders utilize a two-layer architecture to project heterogeneous features into a shared latent space $\mathbb{R}^{d_{\text{out}}}$ (where $d_{\text{out}}=64$), yielding refined node representations $\mathbf{H}_{\text{Ab}}$ and $\mathbf{H}_{\text{Ag}}$. These residue-level features are aggregated into global graph representations, $\mathbf{h}_{\text{Ab}}$ and $\mathbf{h}_{\text{Ag}}$, via global mean pooling. The final embedding for the pair $\mathcal{X}$ is formed by concatenating these vectors: $\mathbf{e} = [\mathbf{h}_{\text{Ab}} \, \| \, \mathbf{h}_{\text{Ag}}] \in \mathbb{R}^{2d_{\text{out}}}$.

\paragraph{PU Pre-training} \label{PU Pre-training}
To enhance representation learning, we employ a PU pre-training strategy (\cref{fig2} (a)) on the combined dataset $\mathcal{D} = \mathcal{D}_L \cup \mathcal{D}_U$, where $\mathcal{D}_L$ and $\mathcal{D}_U$ denote the labeled and unlabeled Ab-Ag pairs, respectively. As a prerequisite, we also generate two augmented views via stochastic graph perturbations: a weakly augmented representation $\mathbf{e}'$ and a strongly augmented representation $\mathbf{e}''$ (see Appendix \cref{Appendix-GNN} for details). Given the continuous affinity $y \in \mathbb{R}$, we first discretize it into a categorical label $y^{\text{cls}} \in \{-1, 0, 1\}$, representing high, medium, and low affinity levels, respectively. To facilitate a unified training objective, we define a target variable $\tilde{y}^{\text{cls}}$ for each Ab-Ag pair. For labeled data ($\mathcal{X} \in \mathcal{D}_L$), we set $\tilde{y}^{\text{cls}} = y^{\text{cls}}$, using the ground truth; for unlabeled data ($\mathcal{X} \in \mathcal{D}_U$), $\tilde{y}^{\text{cls}}$ is the pseudo-label initialized as 0 and refined by the model's prediction.

% \textbf{Dual-Level Contrastive Representation.}
We optimize a multi-task objective on the combined dataset, integrating supervised classification with self-supervised contrastive learning. The classification objective utilizes the weakly augmented view $\mathbf{e}'_i$ and a classifier $f_{\text{cls}}$ to predict the affinity class probabilities $\hat{\mathbf{y}}_i^{cls} = f_{\text{cls}}(\mathbf{e}'_i)$ and the cross-entropy loss ($\mathcal{L}_{\text{CE}}$) is used to minimize the discrepancy between predictions and the targets. Simultaneously, we minimize an instance-level contrastive loss to align the weak and strong views:
\begin{equation}
\begin{split}
    \mathcal{L}_{\text{Ins}} &= - \frac{1}{B} \sum_{i=1}^{B} \log \frac{e^{\mathbf{e}'_i \cdot \mathbf{e}''_i / \tau}}{\Omega_i}, \\
    \text{where} \quad \Omega_i &= \sum_{j \neq i} \left( e^{\mathbf{e}'_i \cdot \mathbf{e}''_j / \tau} + e^{\mathbf{e}'_i \cdot \mathbf{e}'_j / \tau} \right).
\end{split}
\end{equation}
Following a warm-up period, we refine the representation space to group pairs with similar binding patterns by incorporating a cluster-aware loss $\mathcal{L}_{\text{Clus}}$. For each query $\mathbf{e}'_i$, we construct a robust positive set $\mathcal{P}_i$ by integrating the strategies of consensus verification and geometric preservation (refer to \cref{Positive set} for details) \cite{long2024positive}. This loss encourages the query to align with its semantic neighbors: 
% : cluster alignment ($\mathcal{S}_{\text{cluster}}$), prediction consistency ($\mathcal{S}_{\text{pred}}$), and manifold neighborhood ($\mathcal{S}_{\text{KNN}}$) 
\begin{equation}
    \mathcal{L}_{\text{Clus}} = - \frac{1}{B} \sum_{i=1}^{B} \frac{1}{|\mathcal{P}_i|} \sum_{\mathbf{p} \in \mathcal{P}_i} \log \frac{e^{\mathbf{e}'_i \cdot \mathbf{p} / \tau}}{\Omega_i}.
\end{equation}
% defined as $\mathcal{L}_{\text{CE}} = \frac{1}{B} \sum_{i=1}^B \ell_{\text{CE}}(\hat{\mathbf{y}}_i^{cls}, \tilde{y}^{\text{cls}}_i)$
% , where $\Omega_i$ aggregates negative pairs in the batch of size $B$

% \textbf{Meta-Optimized Label Refinement.}
Subsequently, we adopt a meta-learning strategy \cite{ren2018learning} to refine the initial pseudo-labels assigned to $\mathcal{D}_U$. Let $\mathcal{B}_L$ and $\mathcal{B}_U$ denote the indices of labeled and unlabeled data, respectively. We construct the unified target matrix $\tilde{\mathbf{Y}}_{\text{cls}} \in \mathbb{R}^{B \times N_c}$ (where $N_c=3$) by stacking the current targets $\{\tilde{y}^{\text{cls}}_i\}_{i=1}^B$. We then introduce a learnable perturbation matrix $\mathbf{\Delta} \in \mathbb{R}^{B \times N_c}$ to generate soft targets via $\tilde{\mathbf{Y}}_{\text{cls}} + \mathbf{\Delta}$. The optimization proceeds in a bi-level manner. We first perform a virtual gradient step on the model parameters using the soft targets, and then seek the optimal perturbation $\mathbf{\Delta}^*$ that minimizes the loss on a clean validation batch $\mathcal{D}_{L}^{\text{val}}$. Guided by this optimization, we derive refined discrete pseudo-labels $\tilde{\mathbf{Y}}_{\text{cls}}^{\text{meta}}$. Finally, to stabilize training, we update the targets using an exponential moving average (EMA) with the momentum coefficient of $\beta$:
\begin{equation}
    \tilde{\mathbf{Y}}_{\text{cls}}^{(t+1)} = \beta \tilde{\mathbf{Y}}_{\text{cls}}^{(t)} + (1 - \beta) \tilde{\mathbf{Y}}_{\text{cls}}^{\text{meta}}.
\end{equation}
The detailed derivation of the bi-level optimization is available in Appendix \cref{meta learning}.

\paragraph{List Sampling} \label{Data Sampling}
Following pre-training, we fine-tune the model using a listwise ranking objective. To capture subtle affinity differences, we construct informative ranking lists $\mathcal{T} = \{ \mathcal{X}_1, \dots, \mathcal{X}_K \}$ using a homologous antigen sampling strategy (illustrated in \cref{fig2} (b)). Formally, for a randomly sampled seed pair $\mathcal{X}_{\text{se}}$ with antigen sequence $\mathbf{S}_{\text{Ag}}^{\text{se}}$ and ground-truth affinity $y_{\text{se}}$, we retrieve a candidate pool $\mathcal{P}_{\text{se}}$ from labeled dataset $\mathcal{D}_L$. The pool is restricted to candidates that exhibit high antigen homology with the seed yet significant affinity divergence:
\begin{equation}
\begin{split}
    \mathcal{P}_{\text{se}} = \Big\{ \mathcal{X}_i \in \mathcal{D}_L \;\Big|\; & \text{Sim}(\mathbf{S}_{\text{Ag}}^i, \mathbf{S}_{\text{Ag}}^{\text{se}}) \ge \delta_{\text{seq}} \\
    & \land \left| y_{\text{se}} - y_i \right| > y_{\text{cutoff}} \Big\},
\end{split}
\label{eq:sample}
\end{equation}
where $\text{Sim}(\cdot)$ denotes sequence similarity, $\delta_{\text{seq}}$ is the homology threshold, and $y_{\text{cutoff}}$ enforces a minimum margin to mitigate measurement noise. Finally, $K-1$ pairs are selected from $\mathcal{P}_{\text{se}}$ and combined with $\mathcal{X}_{\text{se}}$ to form the list $\mathcal{T}$. Specific sampling configurations for different data splits and auxiliary strategies for global robustness are detailed in Appendix \cref{Appendix-Data Sample}.

\paragraph{MHSA Ranking} \label{Ranking Module}
To map the structural embeddings to scalar affinity ranking scores, we employ a ranking module based on MHSA mechanism. Specifically, we adopt the Induced Set Attention Block (ISAB) variant \cite{lee2019set} to mediate the information flow, by introducing $M$ learnable inducing points as a reference frame. Formally, we construct the input matrix $\mathbf{E}_{\mathcal{T}} \in \mathbb{R}^{K \times 2d_{\text{out}}}$ by stacking the embeddings of the $K$ pairs in the list $\mathcal{T}$. These embeddings are then processed through $N_r$ layers of ISAB (see Appendix \cref{Appendix-Ranking} for mathematical details). Finally, the context-aware features $\mathbf{Z}^{(L)} \in \mathbb{R}^{K \times d_r}$ are projected to scalar scores via a linear layer:
\begin{equation}
\mathbf{r} = \mathbf{Z}^{(L)} \mathbf{w} + b,
\end{equation}
where $\mathbf{w} \in \mathbb{R}^{d_r}$ and $b \in \mathbb{R}$ are learnable parameters, and $\mathbf{r} \in \mathbb{R}^{K}$ represents the ranking scores for the $K$ pairs.

\begin{table*}[t]
\centering
\caption{\textbf{Performance comparison on the Random Split.} We employ 5-fold cross-validation (mean $\pm$ std) for all methods. Due to computational costs, the methods of FoldX and ANTIPASTI (marked with $^*$) were evaluated on a subsampled dataset (see Appendix \cref{Data Split} for details). Accordingly, AbLWR$^*$ reports our model's performance on this specific subset to ensure a fair comparison. All metrics are in percentage (\%) except for $K\tau$, and higher values indicate better performance. Best results are highlighted in \textbf{bold}.}
\label{tab:random_split}
\begin{small}
\begin{tabular}{llccccc}
\toprule
\textbf{Category} & \textbf{Model} & \textbf{FRA (\%)} & \textbf{$K\tau$} & \textbf{PRA (\%)} & \textbf{PAU (\%)} & \textbf{P@1 (\%)} \\
\midrule
\multirow{2}{*}{Sequence-based} 
 & MINT       & 1.11 $\pm$ 0.09          & 0.07 $\pm$ 0.00         & 53.33 $\pm$ 0.16         & 53.33 $\pm$ 0.16         & 23.12 $\pm$ 0.62         \\
 & AntiBERTy  & 5.09 $\pm$ 0.19          & 0.35 $\pm$ 0.01         & 67.59 $\pm$ 0.43         & 67.59 $\pm$ 0.43         & 39.96 $\pm$ 2.45         \\
\midrule
\multirow{2}{*}{Structure-based} 
 & AbRank     & 5.63 $\pm$ 2.78          & 0.31 $\pm$ 0.11         & 65.40 $\pm$ 5.61         & 65.40 $\pm$ 5.61         & 39.64 $\pm$ 5.65         \\
 & GraphDTA   & 8.98 $\pm$ 1.23          & 0.46 $\pm$ 0.03         & 73.02 $\pm$ 1.68         & 73.02 $\pm$ 1.68         & 46.92 $\pm$ 4.90         \\
\midrule
Complex-based & ANTIPASTI$^*$ & 0.00 $\pm$ 0.00 & -0.07 $\pm$ 0.07 & 46.24 $\pm$ 3.69 & 46.24 $\pm$ 3.69 & 17.96 $\pm$ 4.42 \\
\midrule
\multirow{2}{*}{Traditional} 
 & LambdaMART & 6.36 $\pm$ 0.24          & 0.38 $\pm$ 0.01         & 68.18 $\pm$ 0.64         & 68.79 $\pm$ 0.63         & 39.04 $\pm$ 0.48         \\
 & FoldX$^*$     & 2.04 $\pm$ 1.44 & 0.01 $\pm$ 0.09 & 50.24 $\pm$ 4.36 & 50.27 $\pm$ 4.35 & 17.14 $\pm$ 2.74\\
\midrule
\multirow{2}{*}{\textbf{Ours}} 
 & \textbf{AbLWR} & \textbf{20.74} $\pm$ \textbf{0.45} & \textbf{0.59} $\pm$ \textbf{0.01} & \textbf{79.31} $\pm$ \textbf{0.52} & \textbf{79.49} $\pm$ \textbf{0.51} & \textbf{57.26} $\pm$ \textbf{0.56} \\
 & \textbf{AbLWR$^*$} & \textbf{17.14} $\pm$ \textbf{5.88} & \textbf{0.57} $\pm$ \textbf{0.05} & \textbf{78.16} $\pm$ \textbf{2.54} & \textbf{78.30} $\pm$ \textbf{2.65} & \textbf{53.47} $\pm$ \textbf{5.28} \\
\bottomrule
\end{tabular}
\end{small}
\end{table*}

\subsection{Training} \label{Training}
\paragraph{Dataset and Split.} We curated Ab-Ag pairs from eight public repositories, derived from the compilation by Liu et al. \cite{liu2025abrank} (details in Appendix \cref{Data Collection}). Each entry comprises paired variable heavy and light chains alongside their corresponding antigen sequences. Following the established protocol, binding affinity measurements were standardized to the equilibrium dissociation constant ($K_d$) and log-transformed ($\log K_d$) to address measurement heterogeneity. Subsequently, the dataset was subsampled to 60,620 pairs for training and evaluation. We employed three splitting strategies to rigorously assess model generalization: (1) standard Random Split with 5-fold cross-validation; (2) Ag-based Split to test performance across antigen types; and (3) Ab-based Split using CDR clustering to assess robustness against antibody variations. Detailed splitting protocols are provided in Appendix \cref{Data Split}. Additionally, our case studies utilize the influenza virus dataset from Momont et al. \cite{momont2023pan} and the human interleukin-33 (IL-33) dataset from Wang et al. \cite{wang2025generative} (see Appendix \cref{Appendix-Casestudy}).

\paragraph{Training Objectives.}
We adopt a two-stage optimization strategy. The pre-training phase minimizes a composite objective $\mathcal{L}_{\text{pre}} = \mathcal{L}_{\text{CE}} + \mathcal{L}_{\text{Ins}} + \mathcal{L}_{\text{Clus}}$ to learn robust representations. In the fine-tuning stage, we optimize the ranking performance using the ListMLE loss, which maximizes the likelihood of the correct permutation based on the Plackett-Luce model \cite{xia2008listwise}. Let $\pi = [\pi_1, \dots, \pi_K]$ denote the permutation indices sorted by ground-truth affinity in descending order. The objective is defined as:
\begin{equation}
\label{eq:listmle}
\mathcal{L}_{\text{ListMLE}} = - \sum_{i=1}^{K} \log \left( \frac{\exp(r_{\pi_i})}{\sum_{j=i}^{K} \exp(r_{\pi_j})} \right).
\end{equation}

\section{Experiments}
In this section, we present a comprehensive evaluation of AbLWR across diverse data splitting strategies. We begin with the standard Random Split in \cref{SOTA Compare Random}, followed by more challenging scenarios involving distinct clusters in \cref{Hard AgAb Split}. Furthermore, we conduct ablation studies to verify the contribution of AbLWR's key components in \cref{Ablation}. Finally, we demonstrate the practical applicability of AbLWR in real-world biological settings through case studies in \cref{Cases}.

\subsection{Comparison with SOTA Methods under Random Split} \label{SOTA Compare Random}
\textbf{Baseline Models.} We benchmark AbLWR against a comprehensive suite of SOTA methods. We categorize these baselines into: (1) sequence-based: MINT \cite{ullanat2026learning} and AntiBERTy \cite{ruffolo2021deciphering}; (2) structure-based: GraphDTA \cite{nguyen2021graphdta} and AbRank \cite{liu2025abrank}; (3) complex-based: ANTIPASTI \cite{michalewicz2024antipasti}; and (4) traditional methods, including the physics-based FoldX \cite{schymkowitz2005foldx} and the tree-based LambdaMART \cite{burges2010ranknet}. Detailed training configurations are provided in Appendix \cref{Baseline}.

\textbf{Evaluation Metrics.} We employ five metrics to assess ranking performance: (1) Full Rank Accuracy (FRA), reflecting overall ranking quality; (2) Kendall's $\tau$ ($K\tau$), measuring the ordinal association between predicted and ground-truth values; (3) Pairwise Rank Accuracy (PRA) and (4) Pairwise AUC (PAU), quantifying the percentage of correctly ordered sample pairs; and (5) Precision@1 (P@1), measuring the accuracy of the highest-ranked prediction.

\textbf{Experimental Results.} Table \ref{tab:random_split} summarizes the quantitative comparison results on the Random Split setting. Our proposed method, AbLWR, consistently outperforms all competing baselines across all five evaluation metrics. Specifically, in terms of the challenging FRA metric, AbLWR achieves 20.74$\%$, establishing a significant lead over the second-best performing method, GraphDTA (8.98$\%$). Similarly, for the ranking correlation metric $K\tau$, our model attains 0.59, clearly surpassing the strongest baseline (0.46). It is worth noting that AbLWR also demonstrates superior stability, as evidenced by the relatively low standard deviations. Regarding the computationally intensive methods evaluated on the subsampled dataset, our corresponding model AbLWR$^*$ maintains its dominance with an FRA of 17.14$\%$, significantly outperforming both complex-based and traditional baselines.

\subsection{Generalization on Distinct Antigen and Antibody Clusters} \label{Hard AgAb Split}
Table \ref{tab:other_splits_performance} extends our evaluation to the Ag-based and Ab-based splits, designed to assess model robustness under severe distribution shifts. For consistency, we employ the same baseline methods and evaluation metrics as in \cref{SOTA Compare Random}. In the Ag-based scenario, AbLWR demonstrates strong performance with an FRA of 10.28$\%$, representing a significant lead over GraphDTA (6.02$\%$). This advantage is also evident in other ranking metrics, confirming the model's robustness on distinct antigen distributions.

The Ab-based splits (Cluster 5 and Cluster 1) impose a stricter test, leading to a universal decline in performance metrics. Nevertheless, AbLWR retains its primacy in identifying the most potent binders. Specifically, our method maintains a clear advantage in FRA (5.66$\%$ and 3.58$\%$) and P@1 (46.35$\%$ and 39.21 $\%$). This suggests that AbLWR is precise in prioritizing the top candidates, a critical trait for practical antibody discovery. Finally, comparisons on the subsampled dataset reinforce this trend, with AbLWR$^*$ significantly exceeding both FoldX and ANTIPASTI, particularly in the challenging Ab-based clusters where traditional methods struggle.

\begin{table}[t]
\centering
\caption{Performance comparison on Ag-based and Ab-based Splits. The Ag-based split corresponds to the single most divergent cluster, while the Ab-based split comprises the two most distinct clusters.}
\label{tab:other_splits_performance}
\resizebox{\columnwidth}{!}{%
\begin{tabular}{llccccc}
\toprule
\textbf{Category} & \textbf{Model} & \textbf{FRA (\%)} & \textbf{$K\tau$} & \textbf{PRA (\%)} & \textbf{PAU (\%)} & \textbf{P@1 (\%)} \\
\midrule
\multicolumn{7}{c}{\textbf{Ag-based Split}} \\
\midrule
\multirow{2}{*}{\shortstack{Sequence-\\based}} 
 & MINT       & 0.67 & 0.00 & 49.63 & 49.63 & 21.93 \\
 & AntiBERTy  & 3.33 & 0.31 & 64.98 & 64.98 & 34.50 \\
\midrule
\multirow{2}{*}{\shortstack{Structure-\\based}} 
 & AbRank     & 1.61 & 0.15 & 57.20 & 57.20 & 24.21 \\
 & GraphDTA   & 6.02 & 0.42 & 70.96 & 70.96 & 41.59 \\
\midrule
\shortstack{Complex-\\based} & ANTIPASTI$^*$ & 0.00 & -0.01 & 49.39 & 49.39 & 22.45 \\
\midrule
\multirow{2}{*}{Traditional} 
 & LambdaMART & 2.56 & 0.22 & 60.67 & 60.73 & 28.32 \\
 & FoldX$^*$     & 2.04 & 0.10 & 54.49 & 54.49 & 36.73 \\
\midrule
\multirow{2}{*}{\textbf{Ours}} 
 & \textbf{AbLWR} & \textbf{10.28} & \textbf{0.51} & \textbf{75.25} & \textbf{75.62} & \textbf{50.10} \\
 & \textbf{AbLWR$^*$} & \textbf{12.24} & \textbf{0.54} & \textbf{76.53} & \textbf{77.03} & \textbf{59.18} \\
\midrule
\midrule
\multicolumn{7}{c}{\textbf{Ab-based Split (Cluster 5)}} \\
\midrule
\multirow{2}{*}{\shortstack{Sequence-\\based}} 
 & MINT       & 0.82 & -0.02 & 49.03 & 49.03 & 19.30 \\
 & AntiBERTy  & 1.07 & 0.05 & 52.35 & 52.35 & 23.08 \\
\midrule
\multirow{2}{*}{\shortstack{Structure-\\based}} 
 & AbRank     & 4.15 & 0.35 & 67.68 & 67.68 & 41.43 \\
 & GraphDTA   & 1.34 & 0.08 & 53.75 & 53.75 & 24.12 \\
\midrule
\shortstack{Complex-\\based} & ANTIPASTI$^*$ & 0.00 & 0.02 & 50.82 & 50.82 & 20.41 \\
\midrule
\multirow{2}{*}{Traditional} 
 & LambdaMART & 0.99 & 0.07 & 44.99 & 43.22 & 30.10 \\
 & FoldX$^*$     & 0.00 & 0.04 & 51.84 & 51.84 & 16.33 \\
\midrule
\multirow{2}{*}{\textbf{Ours}} 
 & \textbf{AbLWR} & \textbf{5.66} & \textbf{0.37} & \textbf{68.60} & \textbf{68.62} & \textbf{46.35} \\
 & \textbf{AbLWR$^*$} & \textbf{4.08} & \textbf{0.38} & \textbf{68.98} & \textbf{68.98} & \textbf{44.90} \\
\midrule
\midrule
\multicolumn{7}{c}{\textbf{Ab-based Split (Cluster 1)}} \\
\midrule
\multirow{2}{*}{\shortstack{Sequence-\\based}} 
 & MINT       & 0.74 & -0.03 & 48.56 & 48.56 & 18.86 \\
 & AntiBERTy  & 1.06 & 0.02 & 51.08 & 51.08 & 24.43 \\
\midrule
\multirow{2}{*}{\shortstack{Structure-\\based}} 
 & AbRank     & 2.53 & 0.25 & 62.33 & 62.33 & 38.23 \\
 & GraphDTA   & 1.40 & 0.02 & 51.09 & 51.09 & 22.42 \\
\midrule
\shortstack{Complex-\\based} & ANTIPASTI$^*$ & 0.00 & 0.08 & 54.08 & 54.08 & 24.49 \\
\midrule
\multirow{2}{*}{Traditional} 
 & LambdaMART & 1.25 & 0.09 & 53.37 & 54.48 & 23.59 \\
 & FoldX$^*$     & 2.04 & -0.04 & 47.76 & 47.76 & 18.37 \\
\midrule
\multirow{2}{*}{\textbf{Ours}} 
 & \textbf{AbLWR} & \textbf{3.58} & \textbf{0.25} & \textbf{62.66} & \textbf{62.68} & \textbf{39.21} \\
 & \textbf{AbLWR$^*$} & \textbf{8.16} & \textbf{0.28} & \textbf{64.08} & \textbf{64.08} & \textbf{38.78} \\
\bottomrule
\end{tabular}%
}
\end{table}

\subsection{Ablation Studies} \label{Ablation}
To validate the effectiveness of the proposed components in AbLWR, we conducted a comprehensive ablation study focusing on two key aspects: homologous antigen sampling and model architecture. As detailed in \cref{tab:ablation_study}, we compare the full model against four variants:
(1) \textbf{w/ AbRankSampling}: replaces our sampling strategy with the one used in AbRank \cite{liu2025abrank}, which considers affinity divergence but neglects antigen homology;
(2) \textbf{w/o PU Init.}: removes the PU initialization and relies solely on random initialization;
(3) \textbf{w/o Rank. ListMLE}: reformulates the objective as an absolute affinity regression task using Mean Squared Error (MSE) loss instead of the ranking-based ListMLE;
(4) \textbf{w/o Rank. MHSA}: substitutes the MHSA mechanisms with simple Multilayer Perceptrons (MLPs). The results in \cref{tab:ablation_study} demonstrate that all proposed components contribute significantly to the final performance. Most notably, replacing the ranking loss with MSE regression (\textit{w/o Rank. ListMLE}) causes the most significant performance degradation, with FRA falling from 21.28$\%$ to 3.29$\%$. This underscores that modeling relative order is far more effective than absolute regression for this task. Similarly, removing PU initialization (\textit{w/o PU Init.}) leads to a sharp decline to 5.52$\%$, highlighting the critical role of our initialization strategy in leveraging unlabeled data. Furthermore, our homologous antigen sampling strategy outperforms the baseline sampling (\textit{w/ AbRankSampling}) by a substantial margin (8.73$\%$ in FRA), confirming the necessity of incorporating biological homology into the training process.

\begin{table}[t]
\centering
\caption{Ablation study on the test set. We evaluate the impact of different components and strategies.}
\label{tab:ablation_study}
\resizebox{\columnwidth}{!}{%
\begin{tabular}{lccccc}
\toprule
\textbf{Model} & 
\begin{tabular}{@{}c@{}}\textbf{FRA (\%)} \end{tabular} & 
\begin{tabular}{@{}c@{}}\textbf{$K\tau$} \end{tabular} & 
\begin{tabular}{@{}c@{}}\textbf{PRA (\%)} \end{tabular} & 
\begin{tabular}{@{}c@{}}\textbf{PAU (\%)} \end{tabular} & 
\begin{tabular}{@{}c@{}}\textbf{P@1 (\%)} \end{tabular} \\
\midrule
\textbf{AbLWR (Ours)} & \textbf{21.28} & \textbf{0.60} & \textbf{79.96} & \textbf{80.14} & \textbf{57.73} \\
\midrule
\multicolumn{6}{l}{\textit{Sampling Strategy}} \\
\addlinespace[2pt]
\quad w/ AbRankSampling & 12.55 & 0.52 & 75.73 & 75.90 & 52.92 \\
\midrule
\multicolumn{6}{l}{\textit{Component Ablation}} \\
\addlinespace[2pt]
\quad w/o PU Init. & 5.52 & 0.34 & 66.08 & 66.25 & 41.67 \\
\quad w/o Rank. ListMLE & 3.29 & 0.56 & 59.17 & 59.30 & 31.55 \\
\quad w/o Rank. MHSA & 13.90 & 0.25 & 77.65 & 77.83 & 55.70 \\
\bottomrule
\end{tabular}%
}
\end{table}

\subsection{Case Studies on Influenza Viruses and Human Interleukin-33} 
\label{Cases}
To evaluate the practical effectiveness of AbLWR, we conducted case studies on the influenza virus and human IL-33. For the influenza study, we curated 15 Ab-Ag pairs involving three antibodies and five antigens (detailed in Appendix \cref{tab:neutralization IVA IVB}) \cite{momont2023pan}, which yielded 3,003 unique $K=5$ lists through exhaustive combination. Regarding the human IL-33 study, we analyzed the I7 antibody alongside 19 candidates derived from affinity maturation \cite{wang2025generative}, resulting in a total of 15,504 lists. The ranking performance ($K\tau$) is summarized in Appendix \cref{tab:kendall_tau_casestudy} and the detailed visualizations of high-correlation lists (Kendall's $\tau > 0.4$) are provided in Appendix \cref{sfig5}.

\begin{figure}[htb]
  \vskip 0.2in
  \begin{center}
    \centerline{\includegraphics[width=\columnwidth, trim=0 0 0 0, clip]{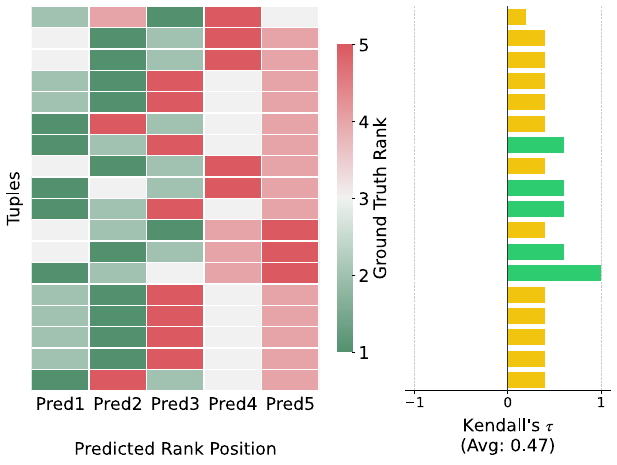}}
    \caption{\textbf{Visualization of ranking predictions on 18 sampled lists.} The heatmap displays the ground truth ranks mapped to the model's predicted positions. Each row represents a test list colored by their ground truth rank, and columns Pred1–Pred5 denote the predicted order. The adjacent bar chart shows the corresponding $K\tau$ calculated for each list row.}
    \label{fig3}
  \end{center}
\end{figure}

\begin{figure}[htb]
  \vskip 0.2in
  \begin{center}
    \centerline{\includegraphics[width=\columnwidth, trim=0 0 0 0, clip]{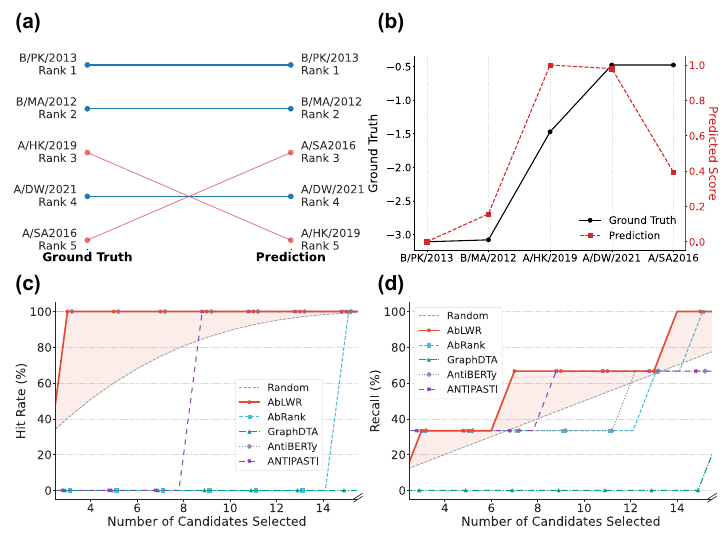}}
    \caption{\textbf{Case studies on FNI17 antibody mechanism and candidates screening for IL-33 antigen.} (a) Comparison of ground truth ranks and predicted ranks. Lines connect the same strain across the two rankings. (b) Dual-axis plot showing the ground truth values and predicted scores for the corresponding viral strains. (c) Hit rate curve for identifying the top-1 binder to the IL-33 antigen. (d) Recall curve showing the percentage of top-3 IL-33 binders retrieved against the number of candidates screened.}
    \label{fig4}
  \end{center}
\end{figure}

We further analyzed the influenza virus dataset from two complementary perspectives. First, we evaluated ranking consistency across 18 test lists derived from the homologous antigen sampling strategy. As shown in \cref{fig3}, the heatmap demonstrates strong alignment: high-affinity pairs (green) are consistently ranked at the top, while low-affinity ones (red) are relegated to the bottom. This observation is supported by the adjacent bar chart, where the majority of tuples exhibit high $K\tau$ scores. Second, we examine FNI17 antibody, which exhibits potent binding to Influenza B (IVB) but significantly reduced affinity for mutated Influenza A (IVA) strains emerging after 2015 \cite{powell2020neuraminidase, wan2019neuraminidase}. As shown in \cref{fig4} (a) and (b), AbLWR effectively discriminates between these viral types. Despite subtle sequence variations, the model correctly predicts the affinity drop in mutated strains, confirming its sensitivity to structural changes driven by viral evolution.
% As shown in \cref{fig4} (a) and (b), AbLWR effectively discriminates between these viral types, confirming its ability to detect subtle affinity changes.

Using the human IL-33 dataset, we demonstrate AbLWR's efficiency in accelerating wet-lab screening. While identifying the optimal binder typically requires exhaustive testing of all 20 candidates, AbLWR significantly streamlines this process. In comparison to the representative data driven baselines, which struggle to surpass random selection in early rounds (\cref{fig4} (c)), AbLWR secures a hit immediately upon the initial selection. Furthermore, it efficiently captures all three best binders within 14 screening rounds (\cref{fig4} (d)). These results demonstrate that AbLWR substantially reduces experimental burden while maintaining high discovery success.
% AbLWR secures a hit immediately upon the initial selection (\cref{fig4} (c)) and efficiently captures all three best binders within 14 screening rounds (\cref{fig4} (d)). 

\section{Conclusions}
We proposed AbLWR, a list-wise framework that redefines binding affinity prediction as a ranking task. By integrating PU learning to mitigate label sparsity and a context-aware architecture to model inter-sample dependencies, AbLWR effectively captures relative binding strengths. Empirical results confirm its superiority over existing SOTA methods and the potential in experimental laboratory settings, such as wet-lab screening. Despite these promising results, the current list sampling strategy is decoupled from the training pipeline and the list size $K$ is a static hyperparameter. In future work, we aim to integrate the sampling process directly into the learning loop and develop adaptive mechanisms to dynamically determine $K$ based on data complexity.

% \section*{Software and Data}
% The AbLWR software is available on Github (...).

% % Acknowledgements should only appear in the accepted version.
% \section*{Acknowledgements}

% \textbf{Do not} include acknowledgements in the initial version of the paper
% submitted for blind review.
\newpage
\section*{Impact Statement}

This paper presents work whose goal is to advance the application of Machine Learning in computational biology. There are many potential societal consequences of our work, none which we feel must be specifically highlighted here.

% % In the unusual situation where you want a paper to appear in the
% % references without citing it in the main text, use \nocite
% \nocite{langley00}

\bibliography{example_paper}
\bibliographystyle{icml2026}

%%%%%%%%%%%%%%%%%%%%%%%%%%%%%%%%%%%%%%%%%%%%%%%%%%%%%%%%%%%%%%%%%%%%%%%%%%%%%%%
%%%%%%%%%%%%%%%%%%%%%%%%%%%%%%%%%%%%%%%%%%%%%%%%%%%%%%%%%%%%%%%%%%%%%%%%%%%%%%%
% APPENDIX
%%%%%%%%%%%%%%%%%%%%%%%%%%%%%%%%%%%%%%%%%%%%%%%%%%%%%%%%%%%%%%%%%%%%%%%%%%%%%%%
%%%%%%%%%%%%%%%%%%%%%%%%%%%%%%%%%%%%%%%%%%%%%%%%%%%%%%%%%%%%%%%%%%%%%%%%%%%%%%%
\newpage
\appendix
\onecolumn
\section{Details for Data Collection and Analysis} \label{Appendix-Dataset}

\subsection{Data Collection} \label{Data Collection}
\begin{figure}[htbp]
  \vskip 0.2in
  \begin{center}
    \centerline{\includegraphics[width=\linewidth, trim=0 0 0 0, clip]{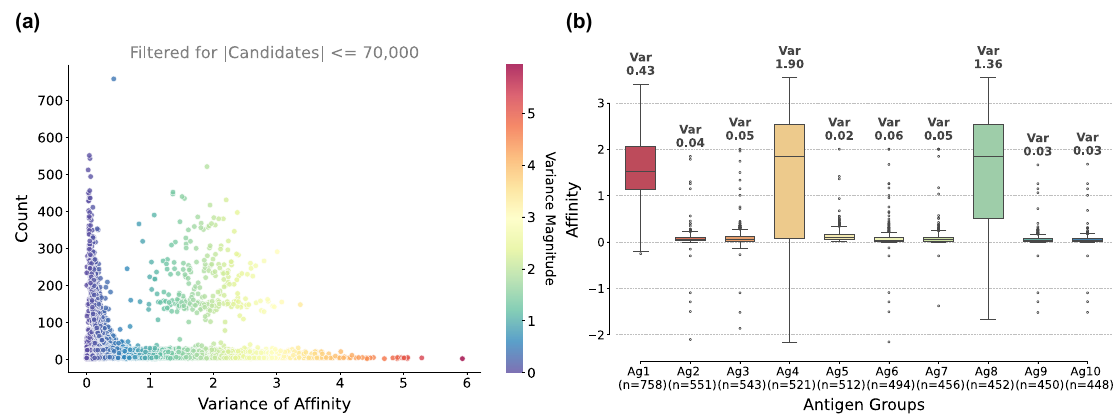}}
    \caption{\textbf{Binding affinity statistics.} (a) Affinity variance per antigen. (b) Affinity distribution for the top 10 most represented antigen groups.}
    \label{sfig1}
  \end{center}
\end{figure}

The distribution of Ab-Ag pairs across different data sources is available from Liu et al. \cite{liu2025abrank}. Brief descriptions are provided below: 
\begin{itemize}
    \item \textbf{RBD-escape \cite{greaney2022antibody}:} Derived from Deep Mutational Scanning (DMS) experiments, this dataset quantifies how mutations in the SARS-CoV-2 Receptor Binding Domain (RBD) affect antibody binding.
    
    \item \textbf{CATNAP \cite{yoon2015catnap}:} A database focused on HIV-1, compiling neutralizing antibody data against HIV-1 Envelope pseudoviruses.
    
    \item \textbf{AlphaSeq \cite{engelhart2022dataset}:} Contains quantitative interaction measurements between designed antibody libraries and specific antigens, generated using the high-throughput AlphaSeq platform.
    
    \item \textbf{AbCoV \cite{rawat2022ab}:} A curated repository specifically for coronaviruses, aggregating binding data for antibodies and nanobodies against SARS-CoV-2, SARS-CoV-1, and MERS-CoV.
    
    \item \textbf{SKEMPI 2.0 \cite{jankauskaite2019skempi}:} A database of thermodynamic binding data ($\Delta \Delta G$ and $K_d$) derived from literature for protein complexes with known 3D structures.
    
    \item \textbf{AbSci \cite{shanehsazzadeh2023unlocking}:} Originates from wet-lab screening campaigns for de novo antibody design, providing binding scores for antibody variants against targets like HER2.
    
    \item \textbf{SAbDab \cite{dunbar2014sabdab}:} The Structural Antibody Database. We utilized the subset of Ab-Ag complexes that have experimentally measured binding affinities annotated in the database.
    
    \item \textbf{OVA-binders \cite{goldstein2019massively}:} Derived from a massively parallel screening study identifying antibody variants that bind to the model antigen Chicken Ovalbumin (OVA).
\end{itemize}
From the initial collection of 342,356 Ab-Ag pairs, we excluded 1,059 entries that lacked binding affinity labels. Furthermore, to address inconsistent annotations, we identified duplicate pairs sharing identical sequences but differing in affinity values. Those with a $\log K_d$ discrepancy exceeding 1.0 were discarded. This filtering process resulted in 340,296 retained pairs, involving 4,750 antigens and 76,017 antibodies as shown in \cref{fig1} (c). Specific statistics information of the selected samples is presented in \cref{sfig1}, demonstrating the significant heterogeneity in binding affinity among different antibodies targeting the same antigen. Subsequently, following the protocol established by Liu et al. \cite{liu2025abrank}, we subsampled the dataset to mitigate bias from dominant antigen groups and verified the validity of all sequences. The final dataset comprises 62,620 Ab-Ag pairs.

% \begin{table}[ht]
% \caption{\textbf{Summary of collected Ab-Ag datasets.} The table lists the source datasets and the number of Ab-Ag pairs collected from each source.}
% \label{ST1}
% \vskip 0.15in
% \begin{center}
% \begin{small}
% \begin{sc}
% \begin{tabular}{lr}
% \toprule
% \textbf{Dataset Source} & \textbf{\# Ab-Ag Pairs} \\
% \midrule
% RBD-escape \cite{greaney2022antibody}      & 192,559 \\
% CATNAP \cite{yoon2015catnap}               & 74,540  \\
% AlphaSeq \cite{engelhart2022dataset}       & 71,834  \\
% AbCoV \cite{rawat2022ab}                   & 1,392   \\
% SKEMPI 2.0 \cite{jankauskaite2019skempi}   & 935     \\
% AbSci \cite{shanehsazzadeh2023unlocking}   & 758     \\
% SAbDab \cite{dunbar2014sabdab}             & 249     \\
% OVA-binders \cite{goldstein2019massively}  & 89      \\
% \midrule
% \textbf{Total}                             & \textbf{342,356} \\
% \bottomrule
% \end{tabular}
% \end{sc}
% \end{small}
% \end{center}
% \vskip -0.1in
% \end{table}

\subsection{Data Split} \label{Data Split}
To comprehensively assess the model’s performance across different scenarios, we designed three distinct data splitting strategies, covering random partitioning and biological sequence segmentation.

\textbf{Random Split.}
Five-fold cross-validation was employed to assess the model’s generalization on randomly split samples, where sequence similarity between training and test sets is not explicitly controlled. 

\begin{figure}[htbp]
  \vskip 0.2in
  \begin{center}
    \centerline{\includegraphics[width=\linewidth, trim=0 0 0 0, clip]{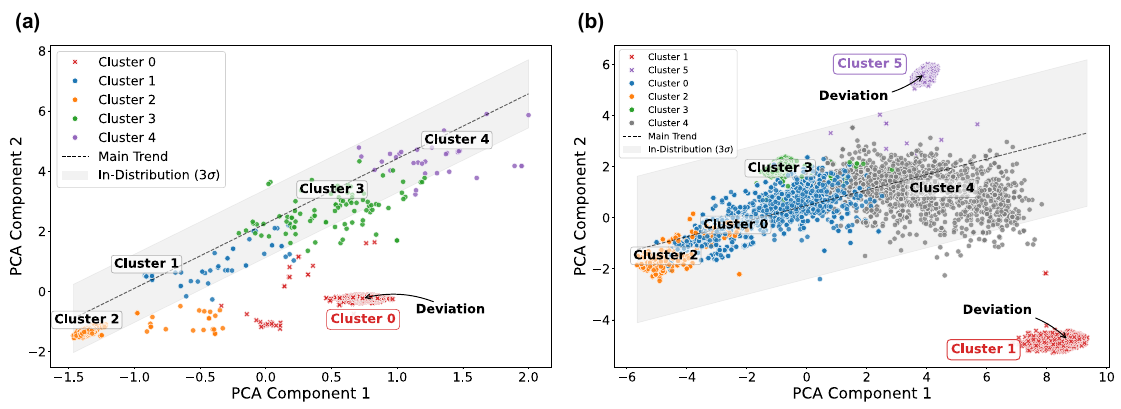}}
    \caption{\textbf{Visualization of sequence clustering.} 
    (a) Antigen sequence clustering. \textit{Cluster 0} is identified as a divergent group distinct from the main distribution and is used to define the Ag-based split. (b) Antibody sequence clustering based on CDRs. \textit{Cluster 1} and \textit{Cluster 5} exhibit significant deviation from the main trend and are selected for the Ab-based split. In both scenarios, these distinct clusters serve as the primary source for the test sets to evaluate model generalization capabilities.}
    \label{sfig2}
  \end{center}
\end{figure}

\textbf{Ag-based Split.}
To evaluate the model's capability of generalizing to novel antigen targets (OOD, Out-of-Distribution generalization), we performed clustering on all antigen sequences (\cref{sfig2} (a)). In real-world antibody engineering, models are often required to predict binding affinities for antigens that differ significantly from the training distribution, such as emerging viral variants or novel protein families \cite{cox2023sars}. To rigorously simulate this distribution shift, we identified the most divergent cluster (Cluster 0), which exhibits the largest spectral distance from the remaining dataset. We allocated 70$\%$ of samples from this cluster to the test set, merging the remaining 30$\%$ with other clusters for training. This partition constructs a few-shot OOD scenario, where the model has limited exposure to the target domain.

\textbf{Ab-based Split.}
To assess the model's ability to prioritize high-affinity pairs among antibodies with distinct paratopes, we clustered antibody sequences based on their CDRs (\cref{sfig2} (b)) and selected the two most divergent clusters (Cluster 5 and Cluster 1). Similar to the antigen split, 70$\%$ of these distinct antibodies were reserved for testing. This split evaluates whether the model can effectively screen for binders that possess low sequence identity to the majority of the training data, a critical requirement for discovering diverse therapeutic candidates.

To facilitate comparison with baseline methods that require explicit binding state prediction, which are computationally intensive, we employed a subsampling strategy. Specifically, we constructed a subset by randomly sampling 50 lists from the original test set in each data split scenario.

\subsection{Data Sampling} \label{Appendix-Data Sample}
% To ensure our model captures the subtle differences required for specific antibody optimization, we primarily employ the \textit{Homologous Antigen Sampling} strategy as detailed in the main text. Additionally, depending on the data split, we strategically incorporate heterogeneous tuples (pairs with divergent antigen sequences) to maintain global ranking robustness.
% \textbf{Homologous Antigen Sampling.}
% The core objective of our fine-tuning phase is to distinguish high-affinity binders from lower-affinity variants within a specific biological context. The candidate pool $\mathcal{P}_{\text{se}}$ is constructed using \cref{eq:sample}. From this pool, we select $K-1$ samples. To maximize the discriminative signal, we prioritize selecting candidates that have the largest affinity margins relative to the seed pair.
We tailor the sampling composition and hyperparameters to the specific objectives of each data split:

\begin{itemize}
\item \textbf{Random and Ag-based Splits:} In these scenarios, the model is required to generalize to diverse antigen structures. Relying solely on homologous sampling could cause the model to degenerate into a set of disjoint local rankers. Therefore, we construct the training data using a balanced 1:1 mixture of homologous lists and heterogeneous lists. These heterogeneous lists, formed by pairs with distinct antigen sequences, force the model to look beyond specific antigen scaffolds and learn universal physicochemical interaction patterns that apply across different biological families. Here, we set the affinity margin threshold $y_{\text{cutoff}}=0.5$.

\item \textbf{Ab-based Split:} For this split, we exclusively employ homologous antigen sampling and increase the threshold to $y_{\text{cutoff}}=1.0$. This design aligns with the antibody lead selection scenario, where the goal is to identify the most potent binders from a diverse pool of candidates targeting the specific antigen (as seen in \cref{fig1} (c)). In the complex antibody space, small affinity differences often arise from noisy or non-generalizable sequence variations. By enforcing a larger margin between all pairs, we provide the model with high-confidence ranking signals, ensuring it learns robust relationships that can transfer to unseen antibody clusters.
% \textcolor{red}{Crucially, given the high diversity of antibody sequences and the challenge of generalizing to unseen antibody clusters, we impose a stricter filtering condition: the affinity difference between any pair of instances $\mathcal{X}_i, \mathcal{X}_j$ within the list must satisfy $|y_i - y_j| > y_{\text{cutoff}}$.} 
\end{itemize}

\section{Details for Ab-Ag GNN Encoders} \label{Appendix-GNN}
\subsection{Graph Construction and Feature Initialization} \label{GNN Initialization}
We independently model the antibody and antigen as graphs $\mathcal{G}=(\mathcal{N}, \mathcal{E})$, where nodes $\mathcal{N}$ represent amino acid residues. The edge set $\mathcal{E}$ is constructed based on spatial proximity derived from the 3D coordinates $\mathbf{X}$. Specifically, an edge $(u, v)$ is established between residue $u$ and residue $v$ if the minimum Euclidean distance between any of their non-hydrogen atoms is less than a threshold of $4.5\,\text{\AA}$\cite{pittala2020learning}. This is formally defined as:
\begin{equation}
    (u, v) \in \mathcal{E} \iff \min_{\mathbf{c} \in \mathbf{x}_u, \mathbf{c}' \in \mathbf{x}_v} \|\mathbf{c} - \mathbf{c}'\|_2 < 4.5,
    \label{eq:edge construction}
\end{equation}
where $\mathbf{x}_u$ and $\mathbf{x}_v$ denote the sets of atomic coordinates for residues $u$ and $v$, respectively. To capture rich semantic information, we initialize node features $\mathbf{H}^{(0)}$ by encoding the primary sequences $\mathbf{S}$ using pre-trained PLMs.
For the antibody, we utilize IgFold \cite{ruffolo2023fast} to encode the CDR sequences:
\begin{equation}
    \mathbf{H}^{(0)}_{\text{Ab}} = \text{IgFold}(\mathbf{S}_{\text{Ab}}) \in \mathbb{R}^{L_{\text{Ab}} \times 512}.
\end{equation}
For the antigen, we employ ESM-2 \cite{lin2023evolutionary} to encode the antigen sequence:
\begin{equation}
    \mathbf{H}^{(0)}_{\text{Ag}} = \text{ESM-2}(\mathbf{S}_{\text{Ag}}) \in \mathbb{R}^{L_{\text{Ag}} \times 1280}.
\end{equation}

\subsection{GNN Encoders} \label{GNN Encoders}
The graph encoders are implemented using the PyTorch Geometric library. As detailed in \cref{eq:edge construction}, the adjacency matrix $\mathbf{A}$ is constructed based on a spatial distance threshold of $4.5\,\text{\AA}$ between residue centroids. We employ \texttt{GCNConv} layers with symmetric normalization. Let $\tilde{\mathbf{A}}$ denote the normalized adjacency matrix. The layer-wise propagation is defined as $\mathbf{H}^{(l+1)} = \sigma ( \tilde{\mathbf{A}} \mathbf{H}^{(l)} \mathbf{W}^{(l)} )$, where $\mathbf{W}^{(l)}$ is the learnable weight matrix and $\sigma(\cdot)$ is the ReLU activation function. The architecture comprises two parallel branches, the antibody encoder and the antigen encoder, both utilizing a two-layer GCN to project high-dimensional features into a shared 64-dimensional latent space. 
% Specific layer-wise hyperparameters are provided in \cref{ST2}.
% \begin{table}[h]
% \centering
% \caption{\textbf{Detailed architecture of the graph encoders.} The output of each GCN layer is followed by a ReLU activation function.}
% \label{ST2}
% \begin{small}
% \begin{tabular}{lcccc}
% \toprule
% \textbf{Module} & \textbf{Layer} & \textbf{Input Dim} & \textbf{Output Dim} & \textbf{Activation} \\
% \midrule
% \multirow{2}{*}{Antibody Encoder ($\phi_{\text{Ab}}$)} & GCN-1 & 512 & 128 & ReLU \\
%  & GCN-2 & 128 & 64 & ReLU \\
% \midrule
% \multirow{2}{*}{Antigen Encoder ($\phi_{\text{Ag}}$)} & GCN-1 & 1280 & 128 & ReLU \\
%  & GCN-2 & 128 & 64 & ReLU \\
% \bottomrule
% \end{tabular}
% \end{small}
% \end{table}

\subsection{Graph Augmentation} \label{GNN Aug}
To construct diverse views required for the PU learning framework, we employ weak and strong augmentations to introduce stochastic perturbations to both structure features and node features.

\textbf{Weak Augmentation.} 
The weak augmentation strategy focuses solely on minor structural perturbations to preserve the semantic integrity of the protein graph while introducing necessary variance. We apply random edge dropout, where edges are removed from the adjacency matrix with a low probability of 0.1.
% $p_{\text{drop}}^{\text{weak}} = 0.1$. Formally, for a graph with edge set $\mathcal{E}$, the augmented edge set $\mathcal{E}'$ is obtained by sampling a subset of edges such that each edge $(u,v) \in \mathcal{E}$ is retained with probability $1 - p_{\text{drop}}^{\text{weak}}$.

\textbf{Strong Augmentation.} 
The strong augmentation strategy applies more aggressive perturbations to both structure and node features to prevent the model from overfitting to specific local patterns. Structural Perturbation: We apply random edge dropout with a higher probability of 0.3. Node Feature Masking: We randomly mask node features to force the model to rely on contextual information with the probability of 0.3. 
% A binary mask $\mathbf{m} \in \{0, 1\}^{|\mathcal{N}|}$ is generated where each entry follows a Bernoulli distribution with probability $1 - p_{\text{mask}}^{\text{strong}}$ (where $p_{\text{mask}}^{\text{strong}} = 0.3$). The node features $\mathbf{X}$ are then updated as $\mathbf{X}' = \mathbf{X} \odot \mathbf{m}$, effectively setting the features of selected residues to zero vectors.

\section{Details for PU Pre-training} \label{Appendix-PUL}
\subsection{Positive set $\mathcal{P}_{i}$ Construction} \label{Positive set}
% To construct a robust positive set $\mathcal{P}_i$ that mitigates the noise from individual views, we integrate global semantic structure, task-specific predictions, and local manifold consistency.

% \textbf{Cluster Alignment ($\mathcal{S}_{\text{cluster}}$).}
% We apply K-means clustering on the representations to assign a cluster prototype $\rho_i$ to each sample $i$ in the training data. The set includes samples sharing the same prototype: $\mathcal{S}_{\text{cluster}}^{(i)} = \{j \mid \rho_j = \rho_i, j \neq i \}$.

% \textbf{Prediction Consistency ($\mathcal{S}_{\text{pred}}$).}
% This set groups samples with identical predicted affinity classes, ensuring task alignment: $\mathcal{S}_{\text{pred}}^{(i)} = \{j \mid \arg\max f_{\text{cls}}(\mathbf{e}'_j) = \arg\max f_{\text{cls}}(\mathbf{e}'_i), j \neq i \}$.

% \textbf{Manifold Neighborhood ($\mathcal{S}_{\text{KNN}}$).}
% To preserve local geometric structure, we select the $k$-nearest neighbors based on cosine similarity: $\mathcal{S}_{\text{KNN}}^{(i)} = \text{Top-}k(\{j \mid j \neq i\}, \mathbf{e}'_i \cdot \mathbf{e}'_j)$.

% Finally, the robust positive set is derived as $\mathcal{P}_i = (\mathcal{S}_{\text{cluster}} \cap \mathcal{S}_{\text{pred}}) \cup \mathcal{S}_{\text{KNN}}$. The intersection term filters noise by requiring consensus between unsupervised clustering and supervised prediction, while the union with $\mathcal{S}_{\text{KNN}}$ ensures the inclusion of local semantic neighbors.

To construct a robust positive set $\mathcal{P}_i$ that mitigates the noise from individual views, we integrate strategies of consensus verification and geometric preservation.

\textbf{Consensus Verification.}
This strategy relies on two complementary candidate sets derived from cluster alignment ($\mathcal{S}_{\text{cluster}}$) and prediction consistency ($\mathcal{S}_{\text{pred}}$). First, we apply K-means clustering on the representations to assign a prototype $\rho_i$ to each sample $i$ and the cluster-based set is defined as $\mathcal{S}_{\text{cluster}}^{(i)} = \{j \mid \rho_j = \rho_i, j \neq i \}$. Second, to ensure task alignment, the prediction consistency set groups samples with identical predicted affinity classes: $\mathcal{S}_{\text{pred}}^{(i)} = \{j \mid \arg\max f_{\text{cls}}(\mathbf{e}'_j) = \arg\max f_{\text{cls}}(\mathbf{e}'_i), j \neq i \}$. Ultimately, only samples present in both sets are incorporated to ensure mutual verification.

\textbf{Geometric Preservation}
To preserve local geometric structure, we select the $k$-nearest neighbors based on cosine similarity: $\mathcal{S}_{\text{KNN}}^{(i)} = \text{Top-}k(\{j \mid j \neq i\}, \mathbf{e}'_i \cdot \mathbf{e}'_j)$.

Finally, the robust positive set is derived by taking the union of the verified consensus and the geometric preservation: $\mathcal{P}_i = (\mathcal{S}_{\text{cluster}} \cap \mathcal{S}_{\text{pred}}) \cup \mathcal{S}_{\text{KNN}}$.

\subsection{Bi-Level Meta-Optimization} \label{meta learning}
The core objective of the meta-optimization is to learn an optimal label perturbation $\mathbf{\Delta}$ for the unlabeled data by leveraging a small, clean validation set $\mathcal{D}_{L}^{\text{val}}$. To strictly preserve experimental ground truth, we enforce $\mathbf{\Delta}_i = \mathbf{0}$ for all $i \in \mathcal{B}_L$. Let $\theta = \{\phi_{\text{Ab}}, \phi_{\text{Ag}}, f_{\text{cls}}\}$ denote the joint parameters of the graph encoders and the classifier. First, we perform a virtual gradient step on the entire classification branch using the soft targets:
\begin{equation}
    \theta'(\mathbf{\Delta}) = \theta - \alpha \nabla_{\theta} \mathcal{L}_{\text{CE}}(f_{\text{cls}}(\mathbf{e}'_{\theta}), \tilde{\mathbf{Y}}_{\text{cls}} + \mathbf{\Delta}),
\end{equation}
where $\alpha$ is the meta-learning rate. Subsequently, we evaluate the virtually updated model $\theta'$ on a clean validation batch $\mathcal{D}_{L}^{\text{val}}$ (with the labels of $\tilde{\mathbf{Y}}_{\text{cls}}^{\text{val}}$) to seek the optimal perturbation $\mathbf{\Delta}^*$ that minimizes the validation loss:
\begin{equation}
    \mathbf{\Delta}^* = \arg\min_{\mathbf{\Delta}} \mathcal{L}_{\text{CE}}(f_{\text{cls}}(\mathbf{e}_{\theta'}), \tilde{\mathbf{Y}}_{\text{cls}}^{\text{val}}).
\end{equation}
Here, $\mathbf{e}_{\theta'}$ indicates the validation representations, computed using the updated encoder parameters within $\theta'$. Guided by this optimization, we derive refined discrete pseudo-labels $\tilde{\mathbf{Y}}_{\text{cls}}^{\text{meta}} = \operatorname{OneHot}(\arg\max(\tilde{\mathbf{Y}}_{\text{cls}} + \mathbf{\Delta}^*))$. Finally, to stabilize training, we update the targets for the subsequent epoch using an Exponential Moving Average (EMA) with the momentum coefficient of $\beta$ \cite{hu2023source}:
\begin{equation}
    \tilde{\mathbf{Y}}_{\text{cls}}^{(t+1)} = \beta \tilde{\mathbf{Y}}_{\text{cls}}^{(t)} + (1 - \beta) \tilde{\mathbf{Y}}_{\text{cls}}^{\text{meta}}.
\end{equation}

The complete bi-level meta-optimization procedure is summarized in \cref{alg:meta_opt} and the pre-training phase is detailed in \cref{alg:pretrain}. The selection for hyperparameters are summarized in \cref{ST3}.

\begin{algorithm}[t]
\caption{Bi-Level Meta-Optimization}
\label{alg:meta_opt}
\begin{algorithmic}[1]
\STATE {\bfseries Input:} Data Batch $\mathcal{B}=\mathcal{B}_{L} \cup \mathcal{B}_{U}$, Validation Data $\mathcal{D}_{L}^{\text{val}}$, Model Parameters $\theta$, Targets $\tilde{\mathbf{Y}}_{\text{cls}}^{(t)}$, Hyperparams $\alpha, \beta$.

\STATE \textbf{Virtual Update:} Compute $\theta'$ with current soft targets:
\STATE $\quad \theta' \leftarrow \theta - \alpha \nabla_{\theta} \mathcal{L}_{\text{CE}}(f_{\text{cls}}(\mathbf{e}_{\theta'}), \tilde{\mathbf{Y}}_{\text{cls}}^{(t)} + \mathbf{\Delta})$

\STATE \textbf{Meta-Optimization:} Find $\mathbf{\Delta}^*$ minimizing validation loss:
\STATE $\quad \mathbf{\Delta}^* \leftarrow \arg\min_{\mathbf{\Delta}} \mathcal{L}_{\text{CE}}(f_{\text{cls}}(\mathbf{e}_{\theta'}), \tilde{\mathbf{Y}}_{\text{cls}}^{\text{val}})$

\STATE \textbf{Update:} Enforce $\mathbf{\Delta}^*_i = \mathbf{0}$ for $i \in \mathcal{B}_L$ and update targets:
\STATE $\quad \tilde{\mathbf{Y}}_{\text{cls}}^{\text{meta}} \leftarrow \operatorname{OneHot}(\arg\max(\tilde{\mathbf{Y}}_{\text{cls}}^{(t)} + \mathbf{\Delta}^*))$
\STATE $\quad \tilde{\mathbf{Y}}_{\text{cls}}^{(t+1)} \leftarrow \beta \tilde{\mathbf{Y}}_{\text{cls}}^{(t)} + (1 - \beta) \tilde{\mathbf{Y}}_{\text{cls}}^{\text{meta}}$

\STATE \textbf{Output:} $\tilde{\mathbf{Y}}_{\text{cls}}^{(t+1)}$
\end{algorithmic}
\end{algorithm}

\begin{algorithm}[tb]
   \caption{PU Pre-training}
   \label{alg:pretrain}
\begin{algorithmic}[1]
   \STATE {\bfseries Input:} Total Dataset $\mathcal{D} = \mathcal{D}_L + \mathcal{D}_U$, Validation Labeled Dataset $\mathcal{D}_L^{\text{val}}$, Encoders $\phi_{\text{Ab}}, \phi_{\text{Ag}}$, Classifier $f_{\text{cls}}$, Warm-up epochs $T_{\text{warm}}$, Momentum $\beta$
   \STATE Initialize parameters $\theta_{\phi}, \theta_{\text{cls}}$, pseudo-labels $\tilde{\mathbf{Y}}_{\text{cls}}$, and cluster-aware loss $\mathcal{L}_{\text{Cluster}} \leftarrow 0$
   
   \FOR{epoch $t = 1$ \textbf{to} MaxEpochs}
      \STATE Sample mini-batch $\mathcal{B} = \mathcal{B}_L \cup \mathcal{B}_U$
      \STATE Generate multi-view representations $\mathbf{e}, \mathbf{e}', \mathbf{e}''$
      
      \STATE $\mathcal{L}_{\text{Ins}} \leftarrow$ Instance Contrastive Loss (Eq. 2)
      
      \IF{$t > T_{\text{warm}}$}
          \STATE \textit{// Bi-Level Meta-Optimization}
          \STATE Refine $\tilde{\mathbf{Y}}_{\text{cls}}$ using \textbf{\cref{alg:meta_opt}}
          
          \STATE \textit{// Cluster-aware Optimization}
          \STATE Identify positive set $\mathcal{P}$ via prototype alignment
          \STATE $\mathcal{L}_{\text{Cluster}} \leftarrow$ Cluster-aware Loss (Eq. 3)
      \ENDIF
      
      \STATE $\mathcal{L}_{\text{CE}} \leftarrow$ Cross-Entropy with targets $\tilde{\mathbf{Y}}_{\text{cls}}$
      \STATE Update $\theta_{\phi}, \theta_{\text{cls}} \leftarrow \text{Optimizer}(\nabla (\mathcal{L}_{\text{CE}} + \mathcal{L}_{\text{Ins}} + \mathcal{L}_{\text{Cluster}}))$
   \ENDFOR
   \STATE \textbf{Output:} Pre-trained encoders $\phi_{\text{Ab}}, \phi_{\text{Ag}}$
\end{algorithmic}
\end{algorithm}

% \section{Details for Data Sampling} \label{Appendix-Data Sample}
% To ensure our model captures the subtle differences required for specific antibody optimization, we primarily employ the \textit{Homologous Antigen Sampling} strategy as detailed in the main text. Additionally, depending on the data split, we strategically incorporate heterogeneous tuples (pairs with divergent antigen sequences) to maintain global ranking robustness.

% \textbf{Homologous Antigen Sampling.}
% As described in \cref{Data Sampling}, the primary objective of our fine-tuning phase is to distinguish high-affinity binders from lower-affinity variants within a specific biological context. The candidate pool $\mathcal{P}_{\text{se}}$ is constructed using \cref{eq:sample}. From this pool, we select $K-1$ samples. To maximize the discriminative signal, we prioritize selecting candidates that have the largest affinity margins relative to the seed pair, ensuring the model learns from hard negatives and clear positives. 

\section{Details for Ranking Module} \label{Appendix-Ranking}
The Ranking Module first projects the input matrix $\mathbf{E}_{\mathcal{T}} \in \mathbb{R}^{K \times 2d_{\text{out}}}$ to a hidden dimension $d_r$, yielding the initial features $\mathbf{Z}^{(0)} \in \mathbb{R}^{K \times d_r}$. Then, $N_r$ layers of ISAB are applied with $M$ learnable inducing points $\mathbf{I} \in \mathbb{R}^{M \times d_r}$ for robust context modeling. The update rule for the $l$-th layer is defined as:
\begin{equation}
\begin{aligned}
    \mathbf{H}_{\text{ind}} &= \operatorname{MHA}(\mathbf{I}, \mathbf{Z}^{(l-1)}, \mathbf{Z}^{(l-1)}) + \mathbf{I}, \\
    \mathbf{Z}^{(l)} &= \operatorname{rFF}(\operatorname{MHA}(\mathbf{Z}^{(l-1)}, \mathbf{H}_{\text{ind}}, \mathbf{H}_{\text{ind}}) + \mathbf{Z}^{(l-1)}),
\end{aligned}
\end{equation}
where $\operatorname{MHA}(\cdot)$ denotes Multi-Head Attention and $\operatorname{rFF}(\cdot)$ is a row-wise feed-forward network. The inducing points $\mathbf{I}$ serve as a global workspace that aggregates information from the input set $\mathbf{Z}^{(l-1)}$ into $\mathbf{H}_{\text{ind}}$, and then distributes the refined context back to update the features \cite{lee2019set}. Detailed implementations of the ranking module and hyperparameter configurations are provided in \cref{alg:ranking} and \cref{ST3}, respectively.

% Since lower binding affinity values signify stronger interactions, we negate both the ground truth $\mathbf{y}$ and predicted scores $\mathbf{r}$ to align with standard ranking conventions. This transformation is applied prior to computing the ListMLE loss (\cref{eq:listmle}). 
\begin{algorithm}[tb]
   \caption{MHSA Ranking}
   \label{alg:ranking}
\begin{algorithmic}[1]
   \STATE {\bfseries Input:} Tuple set $\mathcal{T}$ (constructed via Homologous Sampling)
   \STATE {\bfseries Input:} Pre-trained Encoders $\phi_{\text{Ab}}, \phi_{\text{Ag}}$, Set Transformer $f_{\text{rank}}$
   \STATE {\bfseries Hyperparameters:} Learning rate $\eta$, Batch size $B$
   
   \STATE Initialize ranking module parameters $\theta_{\text{rank}}$
   
   \FOR{epoch $t = 1$ \textbf{to} MaxEpochs}
      \FOR{each mini-batch of tuples $\{\mathcal{T}\}_B$}
          \STATE \textit{// Forward pass}
          \STATE Extract embeddings $\mathbf{E}_{\mathcal{T}}$ for tuples via $\phi_{\text{Ab}}, \phi_{\text{Ag}}$
          \STATE Predict ranking scores $\mathbf{r} = f_{\text{rank}}(\mathbf{E}_{\mathcal{T}})$
          
          \STATE \textit{// Optimization}
          \STATE Compute $\mathcal{L}_{\text{ListMLE}}$ (Eq. \ref{eq:listmle})
          \STATE Update $\theta_{\phi}, \theta_{\text{rank}} \leftarrow \text{Optimizer}(\nabla \mathcal{L}_{\text{ListMLE}})$
      \ENDFOR
   \ENDFOR
\end{algorithmic}
\end{algorithm}

\begin{table}[h]
\caption{Hyperparameter settings for AbLWR. The model is trained in two stages: PU Pre-training (left) and MHSA Ranking (right).}
\label{ST3}
\begin{center}
\begin{small}
\begin{tabular}{lr c lr}
\toprule
\multicolumn{2}{c}{\textbf{PU Pre-training}} & \phantom{a} & \multicolumn{2}{c}{\textbf{MHSA Ranking}} \\
\cmidrule(r){1-2} \cmidrule(l){4-5}
\textbf{Parameter} & \textbf{Value} & & \textbf{Parameter} & \textbf{Value} \\
\midrule
Optimizer & SGD & & Optimizer & Adam \\
Max Epochs & 400 & & Max Epochs & 50 \\
Warm-up Epochs & 20 & & Batch Size & 16 \\
Batch Size & 64 & & Learning Rate & 0.001 \\
Learning Rate & 0.001 & & Weight Decay & 0.3 \\
Meta LR & 0.001 & & Inducing Points ($M$) & 5 \\
Weight Decay & $1 \times 10^{-4}$ & & Attention Heads & 4 \\
Momentum & 0.9 & & Dropout Rate & 0.2 \\
Temperature ($\tau$) & 0.07 & & - & - \\
$\rho$ Range & $[0.8, 0.95]$ & & - & - \\
\bottomrule
\end{tabular}
\end{small}
\end{center}
\end{table}

\section{Training Details} \label{Appendix - Training Details}
\subsection{Baseline Models} \label{Baseline}
We benchmark AbLWR against seven representative baseline methods. Specifically, AbRank adopts a pairwise approach, explicitly learning to contrast the affinity of two Ab-Ag pairs. Conversely, GraphDTA, MINT, AntiBERTy, FoldX, ANTIPASTI, and LambdaMART functions are pointwise regression models, aiming to predict or simulate the absolute binding affinity values for independent Ab-Ag pair. 
% To ensure a fair evaluation, we adapted these methods by aggregating their individual predictions or pairwise preferences to reconstruct the ranking order within each list. For all baselines requiring training, we utilized the same dataset splits (Random, Ag-based, and Ab-based Split) as AbLWR. Specifically, the training data was preprocessed by decomposing each list into individual Ab-Ag pairs and removing duplicates to fit the input requirements of pointwise or pairwise models.

\begin{itemize}
    \item \textbf{AbRank \cite{liu2025abrank}:} A metric-learning framework for Ab-Ag affinity ranking. To adapt it for our specific evaluation, we retrained the model on the datasets provided by the authors, strictly removing any test set leakage (\textit{balanced-train-swapped.csv} for Random Split, \textit{hard-ag-train-swapped.csv} for Ag-based Split, and \textit{hard-ab-train-swapped.csv} for Ab-based Split). We utilized the official implementation from GitHub\footnote{\url{https://github.com/biochunan/AbRank-WALLE-Affinity}} with the standard configuration file \textit{train-abrank.yaml} for all experiments.
    % Additionally, for the test set, we decomposed each list into constituent pairs to evaluate pairwise accuracy. 

    \item \textbf{GraphDTA \cite{nguyen2021graphdta}:} A GNN-based model originally designed to represent molecular interactions as graphs for binding affinity prediction. In our experiments, we adjusted the maximum epochs to 50 and the batch size to 64 due to computational constraints, while keeping all other hyperparameters consistent with the original paper. We adopted the architecture directly from its official repository\footnote{\url{https://github.com/thinng/GraphDTA}}.

    \item \textbf{MINT \cite{ullanat2026learning}:} A PLM-based method designed to model protein interactions in a contextualized manner. We employed MINT as a feature extractor to obtain embeddings for Ab-Ag pairs, followed by a simple MLP to predict binding affinity. We used the pre-trained checkpoint \textit{bernett\_mlp.pth} available on Hugging Face\footnote{\url{https://huggingface.co/varunullanat2012/mint}} with the configuration file \textit{esm2\_t33\_650M\_UR50D.json}. The code was adapted from the official repository\footnote{\url{https://github.com/VarunUllanat/mint}}. 

    \item \textbf{AntiBERTy \cite{ruffolo2021deciphering}:} An antibody-specific transformer pre-trained on 558M natural antibody sequences. Similar to our MINT setup, we utilized the \texttt{AntiBERTyRunner} library to extract embeddings for Ab-Ag pairs, which were then fed into an MLP for binding affinity prediction optimized by MSE loss. We set the hyperparameters (Max Epochs=50, Learning Rate=$1 \times 10^{-3}$) to match our proposed method where applicable. The implementation was based on the official repository\footnote{\url{https://github.com/jeffreyruffolo/AntiBERTy}}.

    \item \textbf{LambdaMART \cite{burges2010ranknet}:} A classic Learning-to-Rank algorithm based on gradient boosted decision trees. We implemented the model using the LightGBM library with the \texttt{lambdarank} objective to directly optimize Mean Reciprocal Rank (MRR) and used AntiBERTy embeddings as input features. The model was trained for 100 boosting rounds with a learning rate of $1 \times 10^{-3}$, a maximum depth of 4, and L1/L2 regularization of 0.1.

    \item \textbf{FoldX \cite{schymkowitz2005foldx}:} A physics-based method that estimates protein stability ($\Delta\Delta G$) based on 3D structural information. Since our dataset lacks experimental complex structures, we utilized HDOCK \cite{yan2020hdock}, a protein-protein docking web server, to generate the binding conformations for each Ab-Ag pair before computing $\Delta\Delta G$.

    \item \textbf{ANTIPASTI \cite{michalewicz2024antipasti}:} A CNN-based model that utilizes normal mode correlation maps derived from elastic network models to predict affinity. Similar to the FoldX setup, we prepared the binding modes using HDOCK prior to prediction. Our implementation followed the tutorial \textit{[Tutorial] Predicting affinity using ANTIPASTI.ipynb} provided in the official GitHub repository\footnote{\url{https://github.com/kevinmicha/ANTIPASTI}}.
\end{itemize}

\subsection{Evaluation Metrics} \label{Metric}
In our experiments, we evaluate the performance on a test set consisting of $N$ query lists. For the $i$-th list $\mathcal{T}^{(i)}$, let $\mathbf{y}^{(i)} = [y_1^{(i)}, \dots, y_K^{(i)}]^\top$ denote the ground-truth binding affinities, and $\mathbf{r}^{(i)} = [r_1^{(i)}, \dots, r_K^{(i)}]^\top$ denote the predicted ranking scores, where $K=5$. The metrics are calculated for each list and then averaged across the entire test set. The detailed definitions of the five metrics are as follows:
\begin{itemize}
    \item \textbf{FRA:} Measuring the percentage of lists where the predicted ranking order perfectly matches the ground-truth order.
    \[
    \text{FRA} = \frac{1}{N} \sum_{i=1}^{N} \mathbb{I}\left(\text{rank}(\mathbf{r}^{(i)}) = \text{rank}(\mathbf{y}^{(i)})\right)
    \]
    where $\mathbb{I}(\cdot)$ is the indicator function, and $\text{rank}(\mathbf{v})$ returns the permutation indices that sort the vector $\mathbf{v}$.

    \item \textbf{$K\tau$:} Evaluating the ordinal association between the predicted score vector $\mathbf{r}$ and the ground-truth vector $\mathbf{y}$.
    \[
    \tau = \frac{1}{N} \sum_{i=1}^{N} \frac{CP^{(i)} - DP^{(i)}}{\frac{1}{2}K(K-1)}
    \]
    where $CP$ and $DP$ denote the number of concordant pairs and discordant pairs within the list, respectively. A value of 1 implies a perfect match, while -1 implies a perfectly reversed order.

    \item \textbf{PRA:} Quantifiying the fraction of correctly ordered pairs within a list to measure the model’s ability to satisfy pairwise constraints.
    \[
    \text{PRA} = \frac{1}{N} \sum_{i=1}^{N} \left[ \frac{1}{\binom{K}{2}} \sum_{u=1}^{K-1} \sum_{v=u+1}^{K} \mathbb{I}\left( (r_u^{(i)} - r_v^{(i)})(y_u^{(i)} - y_v^{(i)}) > 0 \right) \right]
    \]
    where $\binom{K}{2}$ denote the number of possible unique pairs within a list of size K.

    \item \textbf{PAU:} Quantifying the fraction of correctly ordered pairs within each list to measure the model’s ability to satisfy pairwise constraints by reformulating the ranking problem as a binary classification task, and computing the Area Under the ROC Curve (AUC) over predicted score and ground-truth preferences.

    \item \textbf{P@1:} Measuring the proportion of lists where the highest-affinity candidate is correctly identified as the top-1 prediction, reflecting the model’s practical utility in antibody discovery.
    \[
    \text{P@1} = \frac{1}{N} \sum_{i=1}^{N} \mathbb{I}\left(\operatorname*{arg\,min}(\mathbf{r}^{(i)}) = \operatorname*{arg\,min}(\mathbf{y}^{(i)})\right)
    \]
\end{itemize}

\section{Details for Case Studies} \label{Appendix-Casestudy}
\subsection{Influenza Viruses}
\begin{table}[t]
\centering
\caption{\textbf{In vitro neutralization IC$_{50}$ values ($\mu$g/mL) and corresponding PDB codes.} Data are adapted from Momont et al.~\cite{momont2023pan}. Lower IC$_{50}$ values indicate higher potency. Entries marked as `$>50$' highlight the loss of neutralization efficacy (immune escape). `--' denotes that the crystal structure is unavailable.}
\label{tab:neutralization IVA IVB}
\begin{small} % ICML 表格通常稍微缩小字体以适应双栏
\begin{tabular}{lll cccccc}
\toprule
& & & \multicolumn{6}{c}{\textbf{In vitro neutralization IC$_{50}$ ($\mu$g/mL)}} \\
\cmidrule(lr){4-9}
& & & \multicolumn{2}{c}{\textbf{FNI9}} & \multicolumn{2}{c}{\textbf{FNI17}} & \multicolumn{2}{c}{\textbf{FNI19}} \\
\cmidrule(lr){4-5} \cmidrule(lr){6-7} \cmidrule(lr){8-9}
\textbf{Strain} & \textbf{Abbr.} & \textbf{Group} & IC$_{50}$ & PDB & IC$_{50}$ & PDB & IC$_{50}$ & PDB \\
\midrule
A/Darwin/6/2021                 & A/DW/2021 & H3N2     & 1.128 & --        & $>50$ & -- & 2.790 & -- \\
A/HongKong/2671/2019            & A/HK/2019 & H3N2     & 0.073 & 8G3P    & 5.068 & -- & 0.190 & 8G40 \\
A/Singapore/INFIMH-16-0019/2016 & A/SA/2016 & H3N2     & 0.162 & --        & $>50$ & -- & 0.916 & -- \\
\addlinespace % 增加一点间距区分不同组别 (H3N2 vs Yamagata)
B/Massachusetts/2/2012          & B/MA/2012 & Yamagata & 0.204 & --        & 0.125 & -- & 0.472 & 8G3Z \\
B/Phuket/3073/2013              & B/PK/2013 & Yamagata & 0.193 & --        & 0.117 & -- & 0.222 & -- \\
\bottomrule
\end{tabular}
\end{small}
\end{table}

\begin{figure}[htb]
  \vskip 0.2in
  \begin{center}
    \centerline{\includegraphics[width=\linewidth, trim=0 0 0 0, clip]{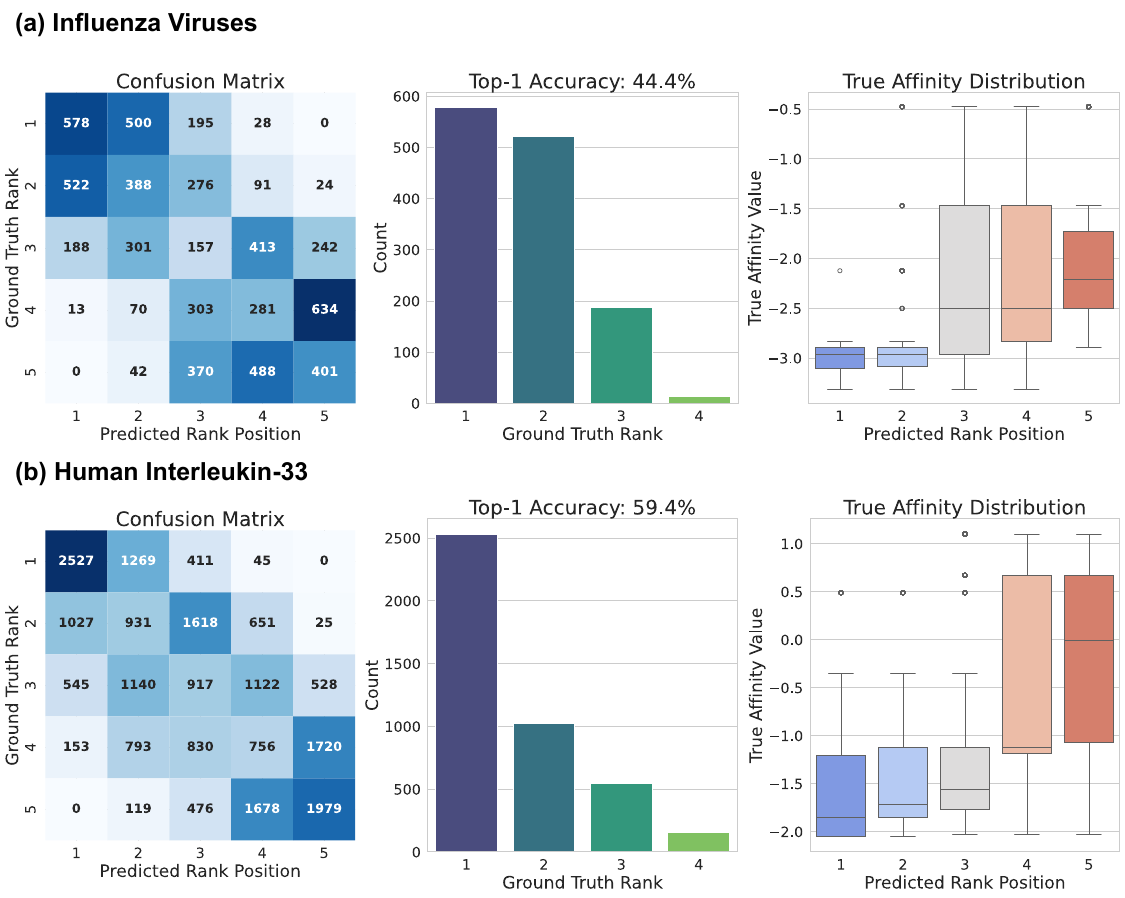}}
    \caption{\textbf{Detailed visualization of ranking performance on (a) influenza viruses and (b) human IL-33.} The comparison includes (left) Confusion Matrices between predicted and ground truth ranks, (middle) Distribution of Ground Truth Ranks for candidates predicted as Top-1, and (right) Boxplots showing the true affinity value distribution across predicted rank positions.}
    \label{sfig5}
  \end{center}
\end{figure}

To validate the practical utility of AbLWR in distinguishing antibody efficacy across evolving viral strains, we selected three monoclonal antibodies (mAbs)—FNI9, FNI19, and FNI17—which have demonstrated broad-spectrum binding against a diverse panel of neuraminidases (NAs) from seasonal and zoonotic influenza A (IAV) and B (IBV) viruses \cite{momont2023pan}. Among these, FNI17 presents a particularly instructive challenge: while it effectively neutralizes historical H3N2 strains, its potency is markedly reduced against variants emerging after 2015 due to antigenic drift. In our analysis, we constructed a specific evaluation set comprising 15 Ab-Ag pairs (involving three antibodies and five antigens, as detailed in \cref{tab:neutralization IVA IVB}). By exhaustively combining these pairs, we included 3,003 unique lists to rigorously test the ranking capabilities of our model. Detailed quantitative results on ranking fidelity are provided in \cref{tab:kendall_tau_casestudy} and \cref{sfig5}, which summarizes the success rates of AbLWR across different Kendall's $\tau$ cutoffs. 

\begin{table}[t]
\centering
\caption{\textbf{Quantitative evaluation of ranking fidelity for AbLWR on influenza virus and human IL-33 datasets.} The values represent the percentage of test cases where the $K\tau$ exceeds the specified thresholds ($0.2, 0.4, 0.6$).}
\label{tab:kendall_tau_casestudy}
\begin{small}
\begin{tabular}{lcc}
\toprule
\textbf{Metric} & \textbf{Influenza Virus} & \textbf{Human IL-33} \\
\midrule
Kendall's $\tau > 0.2$ & 67.83\% & 45.32\% \\
Kendall's $\tau > 0.4$ & 43.32\% & 27.43\% \\
Kendall's $\tau > 0.6$ & 18.02\% & 13.26\% \\
\bottomrule
\end{tabular}
\end{small}
\end{table}

\begin{figure}[htb]
  \vskip 0.2in
  \begin{center}
    \centerline{\includegraphics[width=\linewidth, trim=0 0 0 0, clip]{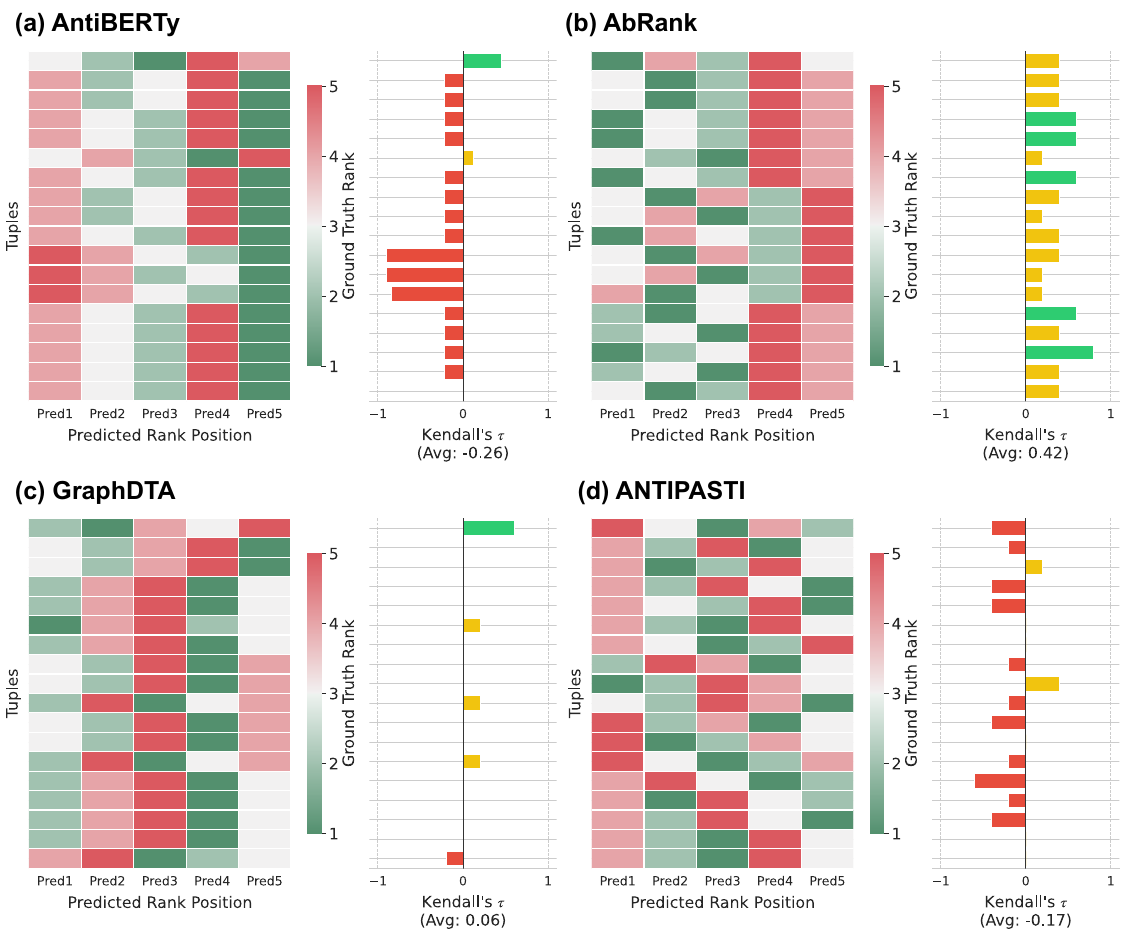}}
    \caption{\textbf{Visualization of ranking predictions on 18 sampled lists by four representative data driven baseline models.} (a) AntiBERTy (sequence-based); (b) AbRank (structure-based pairwise ranking); (c) GraphDTA (structure-based); and (d) ANTIPASTI (complex-based).}
    \label{sfig3}
  \end{center}
\end{figure}

Then, we adopted the same sampling methodology employed for the Ab-based Split (as detailed in \cref{Appendix-Data Sample} with $y_{\text{cutoff}} = 0.1$) to construct the evaluation lists presented in \cref{fig3} (AbLWR) and \cref{sfig3} (representative data driven baselines). As shown in \cref{sfig3}, among these baselines, only AbRank demonstrates a meaningful positive correlation with an average Kendall's $\tau$ of 0.42. Conversely, the other three models—AntiBERTy, GraphDTA, and ANTIPASTI—struggle to distinguish candidates within this narrow affinity range. In comparison, our proposed AbLWR outperforms all baselines, achieving the highest average Kendall's $\tau$ of 0.47. Visually, the heatmap for AbLWR demonstrates a more distinct boundary between high- and low-affinity candidates compared to AbRank.
% Given that the binding affinity values for the Ab-Ag pairs in \cref{fig3} and \cref{sfig3} exhibit a highly compressed dynamic range with subtle variations after unit standardization and log-transformation, we lowered the threshold to $y_{\text{cutoff}} = 0.1$ to capture these fine-grained distinctions. 

\begin{figure}[htb]
  \vskip 0.2in
  \begin{center}
    \centerline{\includegraphics[width=0.95\linewidth, trim=0 0 0 0, clip]{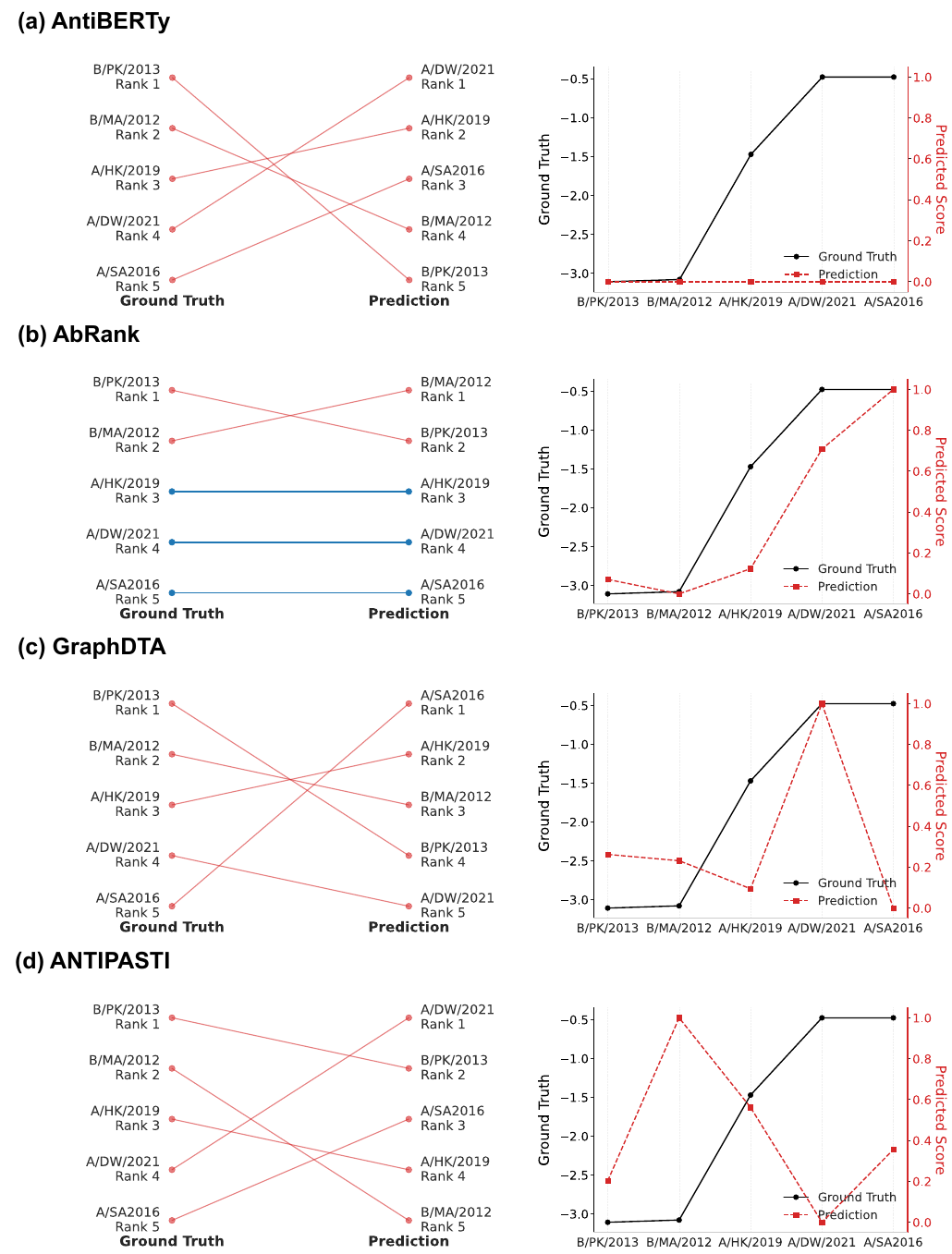}}
    \caption{\textbf{Analysis of FNI17 antibody predictions by four baselines.} Models: (a) AntiBERTy (sequence-based); (b) AbRank (structure-based pairwise ranking); (c) GraphDTA (structure-based); (d) ANTIPASTI (complex-based).}
    \label{sfig4}
  \end{center}
\end{figure}

We further compared the ranking performance on the FNI17 antibody mechanism, as visualized in \cref{sfig4}. Consistent with the results on the sampled lists, AntiBERTy, GraphDTA, and ANTIPASTI failed to capture the correct ranking relationship (lines crossing significantly in \cref{sfig4} (a), (c), (d)). While AbRank (\cref{sfig4} (b)) captures the general ranking trend, this global alignment masks a critical deficiency in practical screening scenarios. As shown in \cref{fig4} (a), AbLWR successfully identified and correctly ordered the top-2 high-affinity strains, placing the ground truth Rank 1 (B/PK/2013) and Rank 2 (B/MA/2012) candidates in their exact predicted positions. In contrast, AbRank failed to distinguish the precise order of these top-tier binders. This demonstrates that our method possesses superior sensitivity in the high-affinity region, which is the most decisive factor for prioritizing lead candidates in antibody discovery.

\subsection{Human IL-33}
\begin{table*}[t]
\caption{\textbf{Experimental $\text{EC}_{50}$ values (nM) of the I7 single-chain antibody and its variants against Human IL-33 \cite{wang2025generative}.} The dataset comprises the wild-type (I7-WT) and 19 generated variants, where lower values indicate stronger binding affinity.}
\label{tab:ec50_values_wide}
\centering
\begin{small}
\begin{tabular}{lrlrlrlr}
\toprule
\textbf{Cand.} & \textbf{$\text{EC}_{50}$} & \textbf{Cand.} & \textbf{$\text{EC}_{50}$} & \textbf{Cand.} & \textbf{$\text{EC}_{50}$} & \textbf{Cand.} & \textbf{$\text{EC}_{50}$} \\
\midrule
I7-WT & 11.41 & I7-M5 & 1898  & I7-M10 & 1.712 & I7-M15 & 705.3 \\
I7-M1 & 2.561 & I7-M6 & 462.2 & I7-M11 & 4.16  & I7-M16 & 1.402 \\
I7-M2 & 37.15 & I7-M7 & 2.898 & I7-M12 & 16.2  & I7-M17 & 1.326 \\
I7-M3 & 148.1 & I7-M8 & 5961  & I7-M13 & 9.857 & I7-M19 & 2.134 \\
I7-M4 & 12.79 & I7-M9 & 9.845 & I7-M14 & 67.34 & I7-M20 & 9.352 \\
\bottomrule
\end{tabular}
\end{small}
\end{table*}

To further evaluate the generalization capability of AbLWR, particularly on antibody architectures distinct from the training distribution, we conducted an additional case study on human Interleukin-33 (IL-33). We utilized the dataset from Wang et al.\cite{wang2025generative}, which describes the optimization of antibody I7 against human IL-33. Crucially, I7 is a single-chain antibody, representing a significant structural deviation from the paired-chain antibodies typically found in standard training datasets. The dataset consists of 20 antibody variants (the wild type and 19 generated variants) with experimentally measured binding affinities (see \cref{tab:ec50_values_wide}). To derive a global ranking for the 20 candidates, we aggregated local predictions from our list-wise model and averaged across all occurrences. The final ranking was determined by sorting these mean scores.

\end{document}